%% file: 0-main.tex
\documentclass[10pt,journal,compsoc]{IEEEtran}

\ifCLASSOPTIONcompsoc
  \usepackage[nocompress]{cite}
  \usepackage{booktabs} 
  \usepackage[group-separator={,}]{siunitx}
  \usepackage{makecell}
  \usepackage[ruled]{algorithm2e}
  \usepackage{amsthm}
  \usepackage{amsmath}
  \usepackage{amsfonts}
  \usepackage{multirow}
  \usepackage{enumitem}
  \usepackage{array}
  \usepackage[skip=3pt]{caption}
  \usepackage{mathrsfs}
  \usepackage{stmaryrd} 
  \usepackage{subfigure}
  \usepackage{balance}
  \usepackage{graphicx}
  \usepackage{amsfonts}
  \usepackage{url}
  \usepackage{xspace}
  \usepackage{upgreek}
  \usepackage{color}
  \usepackage{amsmath}
  \usepackage{makecell}
  \usepackage{multirow}
  \usepackage{array}
  \usepackage{booktabs}
  \usepackage{amssymb}
  \usepackage{microtype}
\else
  \usepackage{cite}
\fi
\usepackage[hidelinks]{hyperref}
\ifCLASSINFOpdf
\else
\fi
\newcommand{\vpara}[1]{\vspace{0.07in}\noindent\textbf{#1 }}

\newcommand{\hide}[1]{} 

\newcolumntype{M}[1]{>{\centering\arraybackslash}m{#1}}

\newcommand{\beq}[1]{\vspace{-0.1in}\begin{equation}#1\end{equation}\vspace{-0.1in}}

\newcommand{\masktoken}[0]{\textsc{[MASK]}\xspace}

\begin{document}
%

\title{\textls[-25]{\spaceskip=0.19em\relax Self-supervised Learning: Generative or Contrastive}}
%
%
%
%

\author{
Xiao Liu, Fanjin Zhang, Zhenyu Hou, Li Mian, Zhaoyu Wang, Jing Zhang, Jie Tang*, \IEEEmembership{IEEE Fellow}
\IEEEcompsocitemizethanks{\IEEEcompsocthanksitem Xiao Liu, Fanjin Zhang, and Zhenyu Hou are with the Department of Computer
Science and Technology, Tsinghua University, Beijing, China.\protect\\
E-mail: liuxiao17@mails.tsinghua.edu.cn, zfj17@mails.tsinghua.edu.cn, hzy17@mails.tsinghua.edu.cn
\IEEEcompsocthanksitem Jie Tang is with the Department of Computer Science and Technology, Tsinghua University, and Tsinghua National Laboratory for Information Science and Technology (TNList), Beijing, China, 100084.\protect\\
E-mail: jietang@tsinghua.edu.cn, corresponding author
\IEEEcompsocthanksitem Li Mian is with the Beijing Institute of Technonlogy, Beijing, China.\protect\\
Email: 1120161659@bit.edu.cn
\IEEEcompsocthanksitem Zhaoyu Wang is with the Anhui University, Anhui, China.\protect\\
Email: wzy950507@163.com
\IEEEcompsocthanksitem Jing Zhang is with the Renmin University of China, Beijing, China.\protect\\
Email: zhang-jing@ruc.edu.cn
}
\thanks{}
}

%
%

\markboth{}%
{Shell \MakeLowercase{\textit{et al.}}: Bare Advanced Demo of IEEEtran.cls for IEEE Computer Society Journals}
%



\IEEEtitleabstractindextext{%
\input{abstract.tex}

}

\maketitle

\IEEEdisplaynontitleabstractindextext

%
\IEEEpeerreviewmaketitle

\tableofcontents
\newpage

\input{1-introduction}
\input{2-motivation}
\input{2.5-overview}
\input{3-generative}
\input{4-contrastive}
\input{5-generative-contrastive}
\input{6-theory}
\input{7-discussion}
\input{7.5-conclusion}
\input{8-acknowledgement}


%

\ifCLASSOPTIONcaptionsoff
  \newpage
\fi



%
\bibliographystyle{abbrv}
\bibliography{ref}
\hide{
}

\input{9-revision}
%

\let\oldaddcontentsline\addcontentsline
\renewcommand{\addcontentsline}[3]{}
\input{bio}
\let\addcontentsline\oldaddcontentsline




\end{document}

%% file: abstract.tex
\begin{abstract}
Deep supervised learning has achieved great success in the last decade. However, its defects of heavy dependence on manual labels and vulnerability to attacks have driven people to find other paradigms. As an alternative, self-supervised learning (SSL) attracts many researchers for its soaring performance on representation learning in the last several years. Self-supervised representation learning leverages input data itself as supervision and benefits almost all types of downstream tasks. In this survey, we take a look into new self-supervised learning methods for representation in computer vision, natural language processing, and graph learning. We comprehensively review the existing empirical methods and summarize them into three main categories according to their objectives: generative, contrastive, and generative-contrastive (adversarial). We further collect related theoretical analysis on self-supervised learning to provide deeper thoughts on why self-supervised learning works. Finally, we briefly discuss open problems and future directions for self-supervised learning. An outline slide for the survey is provided$^1$.
\end{abstract}

%% file: 1-introduction.tex
\section{Introduction}
\IEEEPARstart{D}{}eep neural networks~\cite{lecun2015deep} have shown outstanding performance on various machine learning tasks, especially on supervised learning in computer vision (image classification~\cite{deng2009imagenet,he2016deep,Huang2017Densely}, semantic segmentation~\cite{long2015fully,girshick2014rich}), natural language processing (pre-trained language models~\cite{devlin2019bert,lan2019albert,liu2019roberta,yang2019xlnet}, sentiment analysis~\cite{liu2012sentiment}, question answering~\cite{rajpurkar2016squad,yang2018hotpotqa,ding2019cognitive,asai2019learning} etc.) and graph learning (node classification~\cite{kipf2016semi,qiu2018deepinf,velivckovic2017graph,hu2020heterogeneous}, graph classification~\cite{zhang2018end,bai2019simgnn,sun2019infograph} etc.). Generally, the supervised learning is trained on a specific task with a large labeled dataset, which is randomly divided for training, validation and test.

However, supervised learning is meeting its bottleneck. It relies heavily on expensive manual labeling and suffers from generalization error, spurious correlations, and adversarial attacks. We expect the neural networks to learn more with fewer labels, fewer samples, and fewer trials. As a promising alternative, self-supervised learning has drawn massive attention for its data efficiency and generalization ability, and many state-of-the-art models have been following this paradigm. This survey will take a comprehensive look at the recent developing self-supervised learning models and discuss their theoretical soundness, including frameworks such as Pre-trained Language Models (PTM), Generative Adversarial Networks (GAN), autoencoders and their extensions, Deep Infomax, and Contrastive Coding. An outline slide is also provided.\footnote{Slides at \url{https://www.aminer.cn/pub/5ee8986f91e011e66831c59b/}}

\begin{figure}[t!]
    \centering
    \includegraphics[width=.48\textwidth]{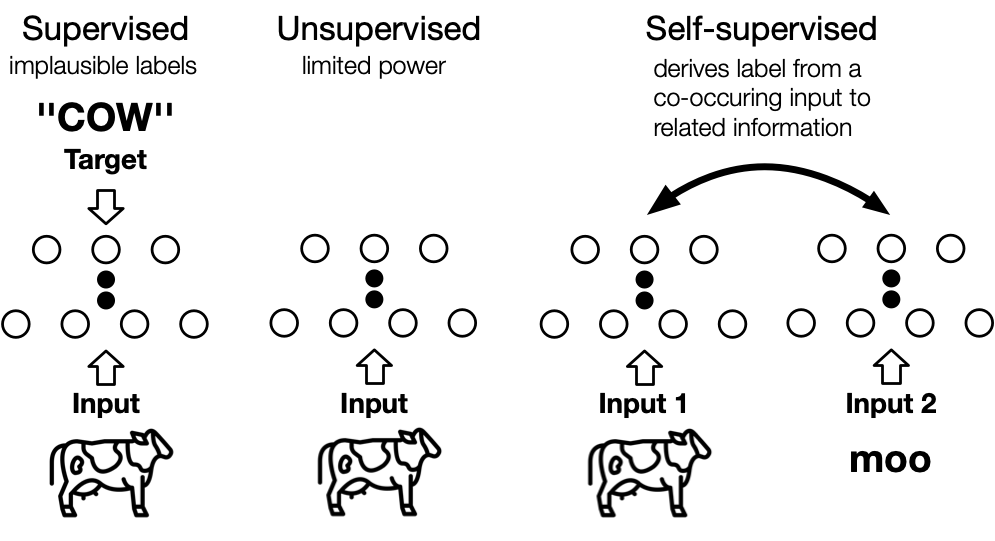}
    \caption{An illustration to distinguish the supervised, unsupervised and self-supervised learning framework. In self-supervised learning, the ``related information'' could be another modality, parts of inputs, or another form of the inputs. Repainted from \cite{de1994learning}.}
    \label{fig:ssl_overview}
    \vspace{-0.4cm}
\end{figure}

\begin{figure}[t!]
    \centering
    \includegraphics[width=.45\textwidth]{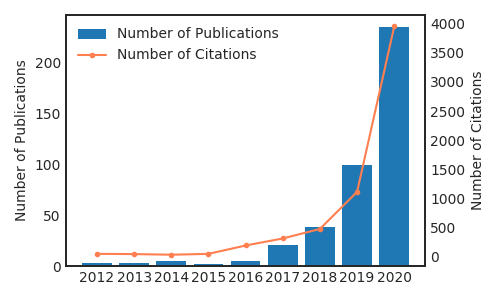}
    \caption{Number of publications and citations on self-supervised learning during 2012-2020, from Microsoft Academic~\cite{sinha2015overview,zhang2019oag}. Self-supervised learning is drawing tremendous attention in recent years.}
    \label{fig:ssl}
    \vspace{-0.4cm}
\end{figure}

The term ``self-supervised learning'' is first introduced in robotics, where training data is automatically labeled by leveraging the relations between different input sensor signals. Afterwards, machine learning community further develops the idea. In the invited speech on AAAI 2020, The Turing award winner Yann LeCun described self-supervised learning as "the machine predicts any parts of its input for any observed part". \footnote{\url{https://aaai.org/Conferences/AAAI-20/invited-speakers/}} Combining self-supervised learning's traditional definition and LeCun's definition, we can further summarize its features as:

\begin{itemize}
    \item Obtain ``labels'' from the data itself by using a ``semi-automatic'' process.
    \item Predict part of the data from other parts.
\end{itemize}

\noindent Specifically, the ``other part'' could be incomplete, transformed, distorted, or corrupted (i.e., data augmentation technique). In other words, the machine learns to 'recover' whole, or parts of, or merely some features of its original input.

People are often confused by the concepts of unsupervised learning and self-supervised learning. Self-supervised learning can be viewed as a branch of unsupervised learning since there is no manual label involved. However, narrowly speaking, unsupervised learning concentrates on detecting specific data patterns, such as clustering, community discovery, or anomaly detection, while self-supervised learning aims at recovering, which is still in the paradigm of supervised settings. Figure \ref{fig:ssl_overview} provides a vivid explanation of the differences between them.

There exist several comprehensive reviews related to Pre-trained Language Models~\cite{qiu2020pre}, Generative Adversarial Networks~\cite{wang2019generative}, autoencoders, and contrastive learning for visual representation~\cite{jing2019self}. However, none of them concentrates on the inspiring idea of self-supervised learning itself. In this work, we collect studies from natural language processing, computer vision, and graph learning in recent years to present an up-to-date and comprehensive retrospective on the frontier of self-supervised learning. To sum up, our contributions are:

\begin{itemize}
    \item We provide a detailed and up-to-date review of self-supervised learning for representation. We introduce the background knowledge, models with variants, and important frameworks. One can easily grasp the frontier ideas of self-supervised learning. 
    \item We categorize self-supervised learning models into generative, contrastive, and generative-contrastive (adversarial), with particular genres inner each one. We demonstrate the pros and cons of each category.
    \item We identify several open problems in this field, analyze the limitations and boundaries, and discuss the future direction for self-supervised representation learning.
\end{itemize}

We organize the survey as follows. In Section \ref{section:motivation}, we introduce the motivation of self-supervised learning. We also present our categorization of self-supervised learning and a conceptual comparison between them. From Section \ref{section:generative} to Section \ref{section:generative-contrastive}, we will introduce the empirical self-supervised learning methods utilizing generative, contrastive and generative-contrastive objectives. 
In Section \ref{section:theory}, we introduce some recent theoretical attempts to understand the hidden mechanism of self-supervised learning's success.
Finally, in Section \ref{section:discussion} and \ref{section:conclusion}, we discuss the open problems, future directions and our conclusions.

%% file: 2-motivation.tex
\section{Motivation of Self-supervised Learning} \label{section:motivation}

\begin{figure}[t]
    \centering
    \includegraphics[width=.48\textwidth]{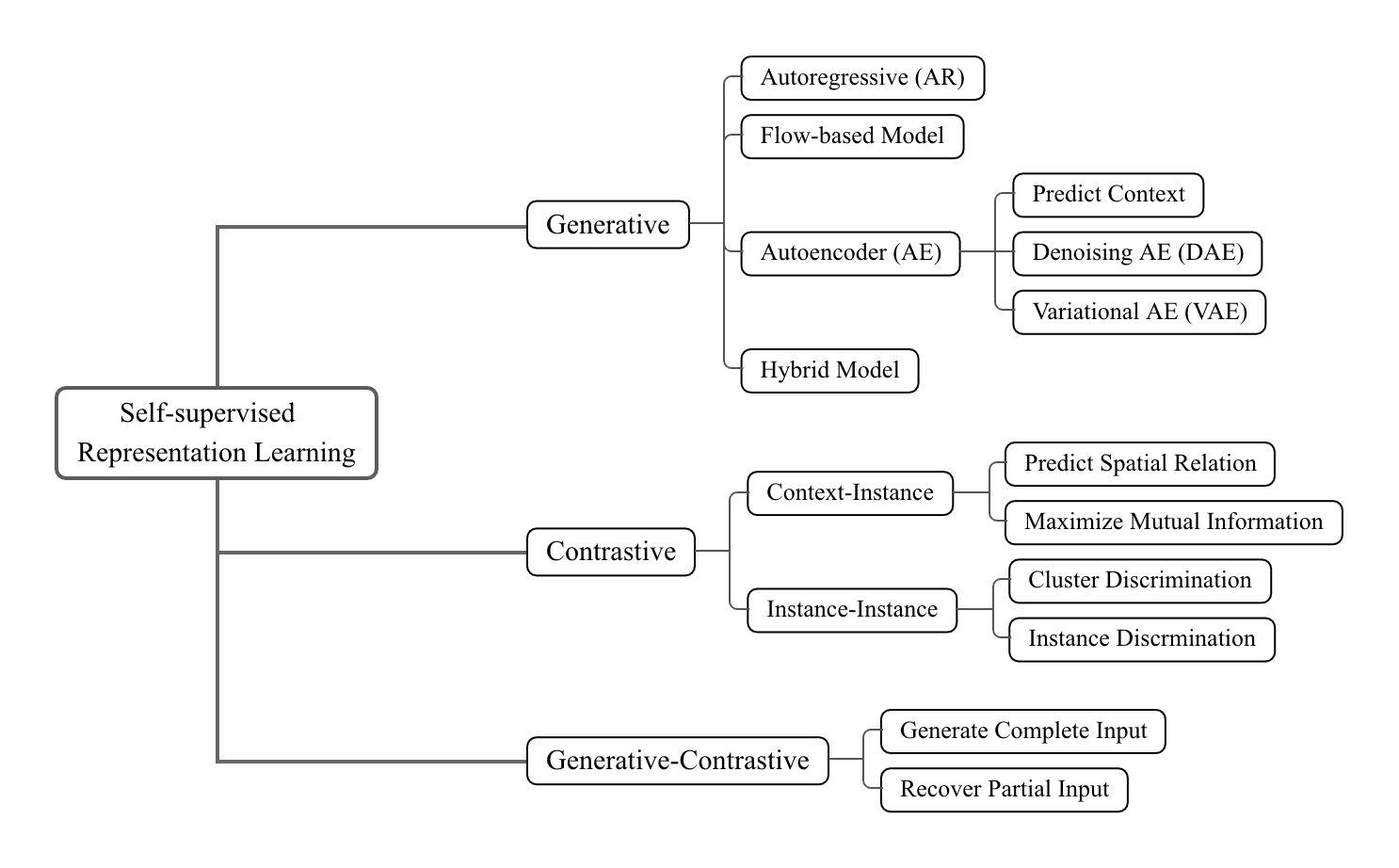}
    \caption{Categorization of Self-supervised learning (SSL): Generative, Contrastive and Generative-Contrastive (Adversarial).}
    \label{fig:category}
    \vspace{-0.2cm}
\end{figure}

It is universally acknowledged that deep learning algorithms are data-hungry. Compared to traditional feature-based methods, deep learning usually follows the so-called ``end-to-end'' fashion (raw-data in, prediction out). It makes very few prior assumptions, which leads to over-fitting and biases in scenarios with little supervised data. Literature has shown that simple multi-layer perceptrons have a very poor generalization ability (always assume a linear relationship for out-of-distribution (OOD) samples)~\cite{xu2020neural}, which results in over-confident (and wrong) predictions. 

To conquer the fundamental OOD and generalization problem, while numerous works focus on designing new architectures for neural networks, another simple yet effective solution is to enlarge the training dataset to make as many samples ``in-distribution''. However, the fact is, despite massive available unlabeled web data in this big data era, high-quality data with human labeling could be costly. For example, Scale.ai\footnote{\url{https://scale.com/pricing}}, a data labeling company, charges \$6.4 per image for the image segmentation labeling. An image segmentation dataset containing 10k+ high-quality samples could cost up to a million-dollar.

The most crucial point for self-supervised learning's success is that it figures out a way to leverage the tremendous amounts of unlabeled data that becomes available in the big data era. It a time for deep learning algorithms to get rid of human supervision and turn back to data's \textit{self-supervision}. The intuition of self-supervised learning is to leverage the data's inherent co-occurrence relationships as the self-supervision, which could be versatile. For example, in the incomplete sentence ``I like \_\_\_\_ apples'', a well-trained language model would predict ``eating'' for the blank (i.e., the famous Cloze Test~\cite{taylor1953cloze}) because it frequently co-occurs with the context in the corpora. We can summarize the mainstream self-supervision into three general categories (see Fig. \ref{fig:category}) and detailed subsidiaries:
\begin{itemize}
    \item Generative: train an encoder to encode input $x$ into an explicit vector $z$ and a decoder to reconstruct $x$ from $z$ (e.g., the cloze test, graph generation)
    \item Contrastive: train an encoder to encode input $x$ into an explicit vector $z$ to measure similarity (e.g., mutual information maximization, instance discrimination)
    \item Generative-Contrastive (Adversarial): train an encoder-decoder to generate fake samples and a discriminator to distinguish them from real samples (e.g., GAN)
\end{itemize}

\begin{figure}[t]
    \centering
    \includegraphics[width=.48\textwidth]{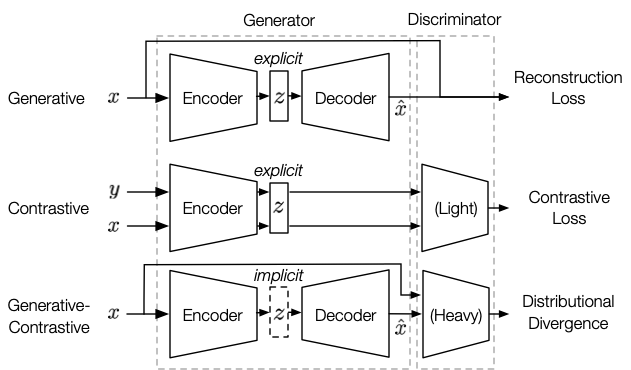}
    \caption{Conceptual comparison between Generative, Contrastive, and Generative-Contrastive methods. }
    \label{fig:ad_three}
    \vspace{-0.2cm}
\end{figure}

Their main difference lies in model architectures and objectives. A detailed conceptual comparison is shown in Fig. \ref{fig:ad_three}. Their architectures can be unified into two general components: the generator and the discriminator, and the generator can be further decomposed into an encoder and a decoder. Different things are:
\begin{enumerate}
    \item For latent distribution $z$: in generative and contrastive methods, $z$ is explicit and is often leveraged by downstream tasks; while in GAN, $z$ is implicitly modeled.
    \item For discriminator: the generative method does not have a discriminator while GAN and contrastive have. Contrastive discriminator has comparatively fewer parameters (e.g., a multi-layer perceptron with 2-3 layers) than GAN (e.g., a standard ResNet~\cite{he2016deep}).
    \item For objectives: the generative methods use a reconstruction loss, the contrastive ones use a contrastive similarity metric (e.g., InfoNCE), and the generative-contrastive ones leverage distributional divergence as the loss (e.g., JS-divergence, Wasserstein Distance).
\end{enumerate}

A properly designed training objective related to downstream tasks could turn our randomly initialized models into excellent pre-trained feature extractors. For example, contrastive learning is found to be useful for almost all visual classification tasks. This is probably because the contrastive object is modeling the class-invariance between different image instances. The contrastive loss makes images containing the same object class more similar. It makes those containing different classes less similar, essentially accords with the downstream image classification, object detection, and other classification-based tasks. The art of self-supervised learning primarily lies in defining proper objectives for unlabeled data.

%% file: 2.5-overview.tex
\begin{table*}[h!]
    \centering
    \scriptsize
    \begin{tabular}{@{\;}c@{\;}|c|c|c|c|c|c|c|c@{\;}}
    \midrule
        Model & FOS & Type & Generator & Self-supervision & Pretext Task & \makecell[c]{Hard\\NS} & \makecell[c]{Hard\\PS} & NS strategy \\ \cline{1-9} 
        \specialrule{0em}{1pt}{0pt}
        GPT/GPT-2~\cite{radford2018improving,radford2019language} & NLP & G & AR & Following words & Next word prediction & - & - & - \\ \cline{1-9} 
        \specialrule{0em}{1pt}{0pt}
        PixelCNN~\cite{van2016conditional,van2016pixel} & CV & G & AR & Following pixels & Next pixel prediction & - & - & - \\ \cline{1-9} 
        \specialrule{0em}{1pt}{0pt}
        NICE~\cite{dinh2014nice} & CV & G & \multirow{3}*{\makecell[c]{Flow\\based}} & \multirow{3}*{Whole image} & \multirow{3}*{Image reconstruction} & - & - & - \\ \cline{1-3}\cline{7-9}
        \specialrule{0em}{1pt}{0pt}
        RealNVP~\cite{dinh2016density} & CV & G & ~ & ~ & ~ & - & - & - \\ \cline{1-3}\cline{7-9}
        \specialrule{0em}{1pt}{0pt}
        Glow~\cite{kingma2018glow} & CV & G & ~ & ~ & ~ & - & - & - \\ \cline{1-9}
        \specialrule{0em}{1pt}{0pt}
        word2vec~\cite{mikolov2013distributed,Mikolov2013EfficientEO} & NLP & G & AE & \multirow{2}*{Context words} & CBOW \& SkipGram & $\times$ & $\times$ & End-to-end \\ \cline{1-4}\cline{6-9}
        \specialrule{0em}{1pt}{0pt}
        FastText~\cite{bojanowski2017enriching} & NLP & G & AE & ~ & CBOW & $\times$ & $\times$ & End-to-end \\ \cline{1-9} 
        \specialrule{0em}{1pt}{0pt}
        \makecell[c]{DeepWalk-based\\~\cite{perozzi2014deepwalk,tang2015line,grover2016node2vec}} & Graph & G & AE & \multirow{2}*{Graph edges} & \multirow{2}*{Link prediction} & $\times$ & $\times$ & End-to-end \\ \cline{1-4}\cline{7-9}
        \specialrule{0em}{1pt}{0pt} 
        VGAE~\cite{kipf2016variational} & Graph & G & AE & ~ & ~ & $\times$ & $\times$ & End-to-end \\ \cline{1-9}
        \specialrule{0em}{1pt}{0pt} 
        BERT~\cite{devlin2019bert} & NLP & G & AE & \makecell[c]{Masked words\\Sentence topic} & \makecell[c]{Masked language model,\\Next senetence prediction} & - & - & - \\ \cline{1-9}
        \specialrule{0em}{1pt}{0pt} 
        SpanBERT~\cite{joshi2020spanbert} & NLP & G & AE & Masked words & Masked language model& - & - & - \\ \cline{1-9}
        \specialrule{0em}{1pt}{0pt} 
        ALBERT~\cite{lan2019albert} & NLP & G & AE & \makecell[c]{Masked words\\Sentence order} & \makecell[c]{Masked language model,\\Sentence order prediction} & - & - & - \\ \cline{1-9}
        \specialrule{0em}{1pt}{0pt} 
        ERNIE~\cite{sun2019ernie,zhang2019ernie} & NLP & G & AE & \makecell[c]{Masked words\\Sentence topic} & \makecell[c]{Masked language model,\\Next senetence prediction} & - & - & - \\ \cline{1-9}
        \specialrule{0em}{1pt}{0pt} 
        GPT-GNN~\cite{hu2020gpt} & Graph & G & AE & \makecell[c]{Attribute \& Edge} & \makecell[c]{Masked graph generation} & - & - & - \\ \cline{1-9}
        \specialrule{0em}{1pt}{0pt} 
        VQ-VAE 2~\cite{razavi2019generating} & CV & G & AE & Whole image & Image reconstruction & - & - & - \\ \cline{1-9}
        \specialrule{0em}{1pt}{0pt} 
        XLNet~\cite{yang2019xlnet} & NLP & G & AE+AR & Masked words & Permutation language model & - & - & - \\ \cline{1-9}
        \specialrule{0em}{1pt}{0pt} 
        GraphAF~\cite{shi2020graphaf} & Graph & G & Flow+AR & Attribute \& Edge & Sequential graph generation & - & - & - \\ \cline{1-9}
        \specialrule{0em}{1pt}{0pt} 
        RelativePosition~\cite{doersch2015unsupervised} & CV & C & - & \multirow{5}*{\makecell[c]{Spatial relations\\(Context-Instance)}} & Relative postion prediction & - & - & - \\ \cline{1-4}\cline{6-9}
        \specialrule{0em}{1pt}{0pt} 
        CDJP~\cite{kim2018learning} & CV & C & - & ~ & \makecell[c]{Jigsaw + Inpainting \\+ Colorization} & $\times$ & $\times$ & End-to-end \\ \cline{1-4}\cline{6-9}
        \specialrule{0em}{1pt}{0pt} 
        PIRL~\cite{misra2019self} & CV & C & - & ~ & Jigsaw & $\times$ & \checkmark & Memory bank \\ \cline{1-4}\cline{6-9}
        \specialrule{0em}{1pt}{0pt} 
        RotNet~\cite{gidaris2018unsupervised} & CV & C & - & ~ & Rotation Prediction & - & - & - \\ \cline{1-9}
        \specialrule{0em}{1pt}{0pt} 
        Deep InfoMax~\cite{hjelm2018learning} & CV & C & - & \multirow{9}*{\makecell[c]{Belonging\\(Context-Instance)}} & \multirow{9}*{MI Maximization} & $\times$ & $\times$ & End-to-end \\ \cline{1-4}\cline{7-9}
        \specialrule{0em}{1pt}{0pt} 
        AMDIM~\cite{bachman2019learning} & CV & C & - & ~ & ~ & $\times$ & \checkmark & End-to-end \\ \cline{1-4}\cline{7-9}
        \specialrule{0em}{1pt}{0pt} 
        CPC~\cite{oord2018representation} & CV & C & - & ~ & ~ & $\times$ & $\times$ & End-to-end \\ \cline{1-4}\cline{7-9}
        \specialrule{0em}{1pt}{0pt} 
        InfoWord~\cite{kong2019mutual} & NLP & C & - & ~ & ~ & $\times$ & $\times$ & End-to-end \\ \cline{1-4}\cline{7-9}
        \specialrule{0em}{1pt}{0pt} 
        DGI~\cite{velivckovic2018deep} & Graph & C & - & ~ & ~ & \checkmark & $\times$ & End-to-end \\ \cline{1-4}\cline{7-9}
        \specialrule{0em}{1pt}{0pt} 
        InfoGraph~\cite{sun2019infograph} & Graph & C & - & ~ & ~ & $\times$ & $\times$ & \makecell[c]{End-to-end\\(batch-wise)} \\ \cline{1-4}\cline{7-9}
        \specialrule{0em}{1pt}{0pt} 
        CMC-Graph~\cite{hassani2020contrastive} & Graph & C & - & ~ & ~ & $\times$ & \checkmark & \makecell[c]{End-to-end} \\ \cline{1-4}\cline{7-9}
        \specialrule{0em}{1pt}{0pt} 
        S$^2$GRL~\cite{Peng2020SelfSupervisedGR} & Graph & C & - & ~ & ~ & $\times$ & $\times$ & End-to-end \\ \cline{1-9}
        \specialrule{0em}{1pt}{0pt} 
        Pre-trained GNN~\cite{hu2019strategies} & Graph & C & - & \makecell[c]{Belonging\\Node attributes} & \makecell[c]{MI maximization,\\Masked attribute prediction} & $\times$ & $\times$ & End-to-end \\ \cline{1-9}               
        \specialrule{0em}{1pt}{0pt} 
        DeepCluster~\cite{caron2018deep} & CV & C & - & \multirow{6}*{\makecell[c]{Similarity\\(Instance-Instance)}} & \multirow{6}*{Cluster discrimination} & - & - & - \\ \cline{1-4}\cline{7-9}
        \specialrule{0em}{1pt}{0pt} 
        Local Aggregation~\cite{zhuang2019local} & CV & C & - & ~ & ~ & - & - & - \\ \cline{1-4}\cline{7-9}
        \specialrule{0em}{1pt}{0pt} 
        ClusterFit~\cite{yan2019clusterfit} & CV & C & - & ~ & ~ & - & - & - \\ \cline{1-4}\cline{7-9} 
        \specialrule{0em}{1pt}{0pt} 
        SwAV~\cite{caron2020unsupervised} & CV & C & - & ~ & ~ & - & \checkmark & \makecell[c]{End-to-end} \\ \cline{1-4}\cline{7-9}
        SEER~\cite{goyal2021selfsupervised} & CV & C & - & ~ & ~ & - & \checkmark & \makecell[c]{End-to-end} \\ \cline{1-4}\cline{7-9}
        M3S~\cite{sun2019multi} & Graph & C & - & ~ & ~ & - & - & - \\ \cline{1-9}     
        \specialrule{0em}{1pt}{0pt} 
        InstDisc~\cite{wu2018unsupervised} & CV & C & - & \multirow{12}*{\makecell[c]{Identity\\(Instance-Instance)}} & \multirow{12}*{Instance discrimination} & $\times$ & $\times$ & Memory bank \\ \cline{1-4}\cline{7-9}
        \specialrule{0em}{1pt}{0pt} 
        CMC~\cite{tian2019contrastive} & CV & C & - & ~ & ~ & $\times$ & \checkmark & End-to-end \\ \cline{1-4}\cline{7-9}
        \specialrule{0em}{1pt}{0pt} 
        MoCo~\cite{he2019momentum} & CV & C & - & ~ & ~ & $\times$ & $\times$ & Momentum \\ \cline{1-4}\cline{7-9}
        \specialrule{0em}{1pt}{0pt} 
        MoCo v2~\cite{chen2020improved} & CV & C & - & ~ & ~ & $\times$ & \checkmark & Momentum \\ \cline{1-4}\cline{7-9} 
        \specialrule{0em}{1pt}{0pt} 
        SimCLR~\cite{chen2020simple} & CV & C & - & ~ & ~ & $\times$ & \checkmark & \makecell[c]{End-to-end} \\ \cline{1-4}\cline{7-9}
        InfoMin~\cite{tian2020makes} & CV & C & - & ~ & ~ & $\times$ & \checkmark & End-to-end \\ \cline{1-4}\cline{7-9}
        BYOL~\cite{grill2020bootstrap} & CV & C & - & ~ & ~ & no NS & \checkmark & \makecell[c]{End-to-end} \\ \cline{1-4}\cline{7-9}
        ReLIC~\cite{mitrovic2020representation} & CV & C & - & ~ & ~ & $\times$ & \checkmark & \makecell[c]{End-to-end} \\ \cline{1-4}\cline{7-9}
        SimSiam~\cite{chen2020exploring} & CV & C & - & ~ & ~ & no NS & \checkmark & \makecell[c]{End-to-end} \\ \cline{1-4}\cline{7-9}
        SimCLR v2 (semi)~\cite{chen2020big} & CV & C & - & ~ & ~ & $\times$ & \checkmark & \makecell[c]{End-to-end} \\ \cline{1-4}\cline{7-9}
        \specialrule{0em}{1pt}{0pt} 
        GCC~\cite{qiu2020gcc} & Graph & C & - & ~ & ~ & $\times$ & \checkmark & Momentum   \\ \cline{1-4}\cline{7-9}
        GraphCL~\cite{you2020graph} & Graph & C & - & ~ & ~ & $\times$ & \checkmark & End-to-end \\ \cline{1-9}     
        \specialrule{0em}{1pt}{0pt} 
        GAN~\cite{goodfellow2014generative} & CV & G-C & AE & \multirow{4}*{Whole image} & \multirow{4}*{Image reconstruction} & - & - & - \\ \cline{1-4}\cline{7-9}
        \specialrule{0em}{1pt}{0pt} 
        Adversarial AE~\cite{makhzani2015adversarial} & CV & G-C & AE & ~ & ~ & - & - & - \\ \cline{1-4}\cline{7-9}
        \specialrule{0em}{1pt}{0pt} 
        BiGAN/ALI~\cite{donahue2016adversarial,dumoulin2016adversarially} & CV & G-C & AE & ~ & ~ & - & - & - \\ \cline{1-4}\cline{7-9}
        \specialrule{0em}{1pt}{0pt} 
        BigBiGAN~\cite{donahue2019large} & CV & G-C & AE & ~ & ~ & - & - & - \\ \cline{1-9}   
        \specialrule{0em}{1pt}{0pt} 
        Colorization~\cite{larsson2016learning} & CV & G-C & AE & Image color & Colorization & - & - & - \\ \cline{1-9}      
        \specialrule{0em}{1pt}{0pt} 
        Inpainting~\cite{pathak2016context} & CV & G-C & AE & Parts of images & Inpainting & - & - & - \\ \cline{1-9}     
        \specialrule{0em}{1pt}{0pt} 
        Super-resolution~\cite{ledig2017photo} & CV & G-C & AE & Details of images & Super-resolution & - & - & - \\ \cline{1-9}      
        \specialrule{0em}{1pt}{0pt} 
        ELECTRA~\cite{clark2020electra} & NLP & G-C & AE & Masked words & Replaced token detection & \checkmark & $\times$ & End-to-end \\ \cline{1-9}   
        \specialrule{0em}{1pt}{0pt} 
        WKLM~\cite{xiong2019pretrained} & NLP & G-C & AE & Masked entities & Replaced entity detection & \checkmark & $\times$ & End-to-end \\ \cline{1-9}     
        \specialrule{0em}{1pt}{0pt} 
        ANE~\cite{dai2018adversarial} & Graph & G-C & AE & \multirow{2}*{Graph edges} & \multirow{2}*{Link prediction} & - & - & - \\ \cline{1-4}\cline{7-9}    
        \specialrule{0em}{1pt}{0pt} 
        GraphGAN~\cite{wang2018graphgan} & Graph & G-C & AE & ~ & ~ & - & - & - \\ \cline{1-9}       
        \specialrule{0em}{1pt}{0pt} 
        GraphSGAN~\cite{ding2018semi} & Graph & G-C & AE & Graph nodes & Node classification & - & - & - \\ \cline{1-9}     
        \specialrule{0em}{1pt}{0pt}
    \end{tabular}
    \caption{An overview of recent self-supervised representation learning. \textmd{For acronyms used, ``FOS'' refers to fields of study; ``NS'' refers to negative samples; ``PS'' refers to positive samples; ``MI'' refers to mutual information. For alphabets in ``Type'': G Generative ; C Contrastive; G-C Generative-Contrastive (Adversarial). For symbols in ``Hard NS'' and ``Hard PS'', ``-'' means not applicable, ``$\times$'' means not adopted, ``\checkmark''' means adopted; ``no NS'' particularly means not using negative samples in instance-instance contrast.}}
    \label{tab:my_label}
\end{table*}

%% file: 3-generative.tex
\section{Generative Self-supervised Learning} \label{section:generative}
This section will introduce important self-supervised learning methods based on generative models, including auto-regressive (AR) models, flow-based models, auto-encoding (AE) models, and hybrid generative models.
\subsection{Auto-regressive (AR) Model}
Auto-regressive (AR) models can be viewed as ``Bayes net structure'' (directed graph model). The joint distribution can be factorized as a product of conditionals

\beq{
\max\limits_{\theta} p_{\theta}(\mathbf{x}) = \sum_{t=1}^{T} \log p_{\theta}(x_t | \mathbf{x}_{1:t-1})
}

\noindent where the probability of each variable is dependent on the previous variables.

In NLP, the objective of auto-regressive language modeling is usually maximizing the likelihood under the forward autoregressive factorization~\cite{yang2019xlnet}.
GPT~\cite{radford2018improving} and GPT-2~\cite{radford2019language} use Transformer decoder architecture~\cite{vaswani2017attention} for language model. 
Different from GPT, GPT-2 removes the fine-tuning processes of different tasks.
To learn unified representations that generalize across different tasks, GPT-2 models $p(output | input, task)$, which means given different tasks, the same inputs can have different outputs. 

The auto-regressive models have also been employed in computer vision, such as PixelRNN~\cite{van2016pixel} and PixelCNN~\cite{van2016conditional}. The general idea is to use auto-regressive methods to model images pixel by pixel. For example, the lower (right) pixels are generated by conditioning on the upper (left) pixels. The pixel distributions of PixelRNN and PixelCNN are modeled by RNN and CNN, respectively. For 2D images, auto-regressive models can only factorize probabilities according to specific directions (such as right and down). Therefore, masked filters are employed in CNN architecture.
Furthermore, two convolutional networks are combined to remove the blind spot in images. Based on PixelCNN, WaveNet~\cite{vanwavenet} -- a generative model for raw audio was proposed. To deal with long-range temporal dependencies, the authors develop dilated causal convolutions to improve the receptive field. Moreover, Gated Residual blocks and skip connections are employed to empower better expressivity.

The auto-regressive models can also be applied to graph domain problems, such as graph generation. You et al.~\cite{you2018graphrnn} propose GraphRNN to generate realistic graphs with deep auto-regressive models. 
They decompose the graph generation process into a sequence generation of nodes and edges conditioned on the graph generated so far. 
The objective of GraphRNN is defined as the likelihood of the observed graph generation sequences.
GraphRNN can be viewed as a hierarchical model, where a graph-level RNN maintains the state of the graph and generates new nodes, while an edge-level RNN generates new edges based on the current graph state.
After that, MRNN~\cite{popova2019molecularrnn} and GCPN~\cite{you2018graph} are proposed as auto-regressive approaches. MRNN and GCPN both use a reinforcement learning framework to generate molecule graphs through optimizing domain-specific rewards. However, MRNN mainly uses RNN-based networks for state representations, but GCPN employs GCN-based encoder networks.

The advantage of auto-regressive models is that they can model the context dependency well. However, one shortcoming of the AR model is that the token at each position can only access its context from one direction.

\subsection{Flow-based Model}

The goal of flow-based models is to estimate complex high-dimensional densities $p(x)$ from data. Intuitively, directly formalizing the densities is difficult. To obtain a complicated densities, we hope to generate it ``step by step'' by stacking a series of transforming functions that describing different data characteristics respectively. Generally, flow-based models first define a latent variable $z$ which follows a known distribution $p_{Z}(z)$. Then define $z = f_{\theta}(x)$, where $f_{\theta}$ is an invertible and differentiable function. The goal is to learn the transformation between $x$ and $z$ so that the density of $x$ can be depicted. According to the integral rule, $p_{\theta} (x) dx = p(z) dz$. Therefore, the densities of $x$ and $z$ satisfies:

\beq{
p_{\theta} (x) = p(f_{\theta}(x)) \bigg| \frac{\partial f_{\theta}(x)}{\partial x}\bigg|
\label{eq:density_trans}
}

\noindent and the objective is to maximize the likelihood:

\begin{align}
\begin{split}
\max\limits_{\theta} \sum\limits_{i} \log p_{\theta} (x^{(i)}) &= \max\limits_{\theta} \sum\limits_{i} \log p_{Z} (f_{\theta}(x^{(i)})) \\
&+ \log \bigg|\frac{\partial f_{\theta}}{\partial x} (x^{(i)}) \bigg|
\end{split}
\end{align}

\noindent The advantage of flow-based models is that the mapping between $x$ and $z$ is invertible. 
However, it also requires that $x$ and $z$ must have the same dimension.
$f_{\theta}$ needs to be carefully designed since it should be invertible and the Jacobian determinant in Eq. (\ref{eq:density_trans}) should also be calculated easily. NICE~\cite{dinh2014nice} and RealNVP~\cite{dinh2016density} design affine coupling layer to parameterize $f_{\theta}$. The core idea is to split $x$ into two blocks $(x_1, x_2)$ and apply a transformation from $(x_1, x_2)$ to $(z_1, z_2)$ in an auto-regressive manner, that is $z_1 = x_1$ and $z_2 = x_2 + m(x_1)$. More recently, Glow~\cite{kingma2018glow} was proposed and it introduces invertible $1 \times 1$ convolutions and simplifies RealNVP.

\subsection{Auto-encoding (AE) Model}
The auto-encoding model's goal is to reconstruct (part of) inputs from (corrupted) inputs. Due to its flexibility, the AE model is probably the most popular generative model with many variants.

\subsubsection{Basic AE Model}
Autoencoder (AE) was first introduced in~\cite{ballard1987modular} for pre-training artificial neural networks. 
Before autoencoder, Restricted Boltzmann Machine (RBM)~\cite{smolensky1986information} can also be viewed as a special ``autoencoder''.
RBM is an undirected graphical model, and it only contains two layers: the visible layer and the hidden layer.
The objective of RBM is to minimize the difference between the marginal distribution of models and data distributions.
In contrast, an autoencoder can be regarded as a directed graphical model, and it can be trained more efficiently.
Autoencoder is typically for dimensionality reduction. Generally, the autoencoder is a feed-forward neural network trained to produce its input at the output layer. The AE is comprised of an \textbf{encoder} network $h = f_{enc} (x)$ and a \textbf{decoder} network $x^{'} = f_{dec} (h)$. The objective of AE is to make $x$ and $x^{'}$ as similar as possible (such as through mean-square error). It can be proved that the linear autoencoder corresponds to the PCA method. Sometimes the number of hidden units is greater than the number of input units, and some interesting structures can be discovered by imposing sparsity constraints on the hidden units~\cite{ng2011sparse}.

\subsubsection{Context Prediction Model (CPM)}

The idea of the Context Prediction Model (CPM) is to predict contextual information based on inputs.

In NLP, when it comes to self-supervised learning on word embedding, CBOW and Skip-Gram~\cite{mikolov2013distributed} are pioneering works. CBOW aims to predict the input tokens based on context tokens. In contrast, Skip-Gram aims to predict context tokens based on input tokens. Usually, negative sampling is employed to ensure computational efficiency and scalability. 
Following CBOW architecture, FastText~\cite{bojanowski2017enriching} is proposed by utilizing subword information.

Inspired by the progress of word embedding models in NLP, many network embedding models are proposed based on a similar context prediction objective. Deepwalk~\cite{perozzi2014deepwalk} samples truncated random walks to learn latent node embedding based on the Skip-Gram model. It treats random walks as the equivalent of sentences. However, another network embedding approach LINE~\cite{tang2015line} aims to generate neighbors rather than nodes on a path based on current nodes:

\beq{
O = -\sum\limits_{(i, j) \in E} w_{ij} \log p(v_j | v_i)
}

\noindent where $E$ denotes edge set, $v$ denotes the node, $w_{ij}$ represents the weight of edge $(v_i, v_j)$. LINE also uses negative sampling to sample multiple negative edges to approximate the objective.

\subsubsection{Denoising AE Model}
The intuition of denoising autoencoder models is that representation should be robust to the introduction of noise.
The masked language model (MLM), one of the most successful architectures in natural language processing, can be regarded as a denoising AE model.
To model text sequence, the masked language model (MLM) randomly masks some of the tokens from the input and then predicts them based on their context information, which is similar to the \textit{Cloze} task~\cite{taylor1953cloze}. BERT~\cite{devlin2019bert} is the most representative work in this field. Specifically, in BERT, a unique token \masktoken is introduced in the training process to mask some tokens. However, one shortcoming of this method is that there are no input \masktoken tokens for down-stream tasks. To mitigate this, the authors do not always replace the predicted tokens with \masktoken in training. Instead, they replace them with original words or random words with a small probability.

Following BERT, many extensions of MLM emerge. SpanBERT~\cite{joshi2020spanbert} chooses to mask continuous random spans rather than random tokens adopted by BERT. Moreover, it trains the span boundary representations to predict the masked spans, inspired by ideas in coreference resolution. ERNIE (Baidu)~\cite{sun2019ernie} masks entities or phrases to learn entity-level and phrase-level knowledge, which obtains good results in Chinese natural language processing tasks. ERNIE (Tsinghua)~\cite{zhang2019ernie} further integrates knowledge (entities and relations) in knowledge graphs into language models.

Compared with the AR model, in denoising AE for language modeling, the predicted tokens have access to contextual information from both sides. However, the fact that MLM assumes the predicted tokens are independent if the unmasked tokens are given (which does not hold in reality) has long been considered as its inherent drawback.

In graph learning, Hu et al.~\cite{hu2020gpt} proposes GPT-GNN, a generative pre-training method for graph neural networks. It also leverages the graph masking techniques and then asks the graph neural network to generate masked edges and attributes. GPT-GNN's wide range of experiments on OAG~\cite{tang2008arnetminer,sinha2015overview,zhang2019oag}, the largest public, academic graph with 100 million nodes and 2 billion edges, shows impressive improvements on various graph learning tasks.

\subsubsection{Variational AE Model}
The variational auto-encoding model assumes that data are generated from underlying latent (unobserved) representation. The posterior distribution over a set of unobserved variables $Z = \{z_1, z_2, ..., z_n\}$ given some data $X$ is approximated by a variational distribution $q(z|x) \approx p(z|x)$. In variational inference, the evidence lower bound (ELBO) on the log-likelihood of data is maximized during training.

\beq{
\log p(x) \geq - D_{KL} (q (z | x) || p(z)) + \mathbb{E}_{\sim q (z | x) } [\log p(x|z)]
}

\noindent where $p(x)$ is evidence probability, $p(z)$ is prior and $p(x|z)$ is likelihood probability. The right-hand side of the above equation is called ELBO. From the auto-encoding perspective, the first term of ELBO is a regularizer forcing the posterior to approximate the prior. The second term is the likelihood of reconstructing the original input data based on latent variables.

Variational Autoencoders (VAE)~\cite{kingma2013auto} is one important example where variational inference is utilized. VAE assumes the prior $p(z)$ and the approximate posterior $q(z|x)$ both follow Gaussian distributions. Specifically, let $p(z) \sim \mathcal{N}(0, 1)$. Furthermore, reparameterization trick is utilized for modeling approximate posterior $q(z|x)$. Assume $z \sim \mathcal{N}(\mu, \sigma^2)$, $z = \mu + \sigma \epsilon$ where $\epsilon \sim \mathcal{N}(0, 1)$. Both $\mu$ and $\sigma$ are parameterized by neural networks. Based on calculated latent variable $z$, decoder network is utilized to reconstruct the input data.

\begin{figure}
    \centering
    \includegraphics[width=.48\textwidth]{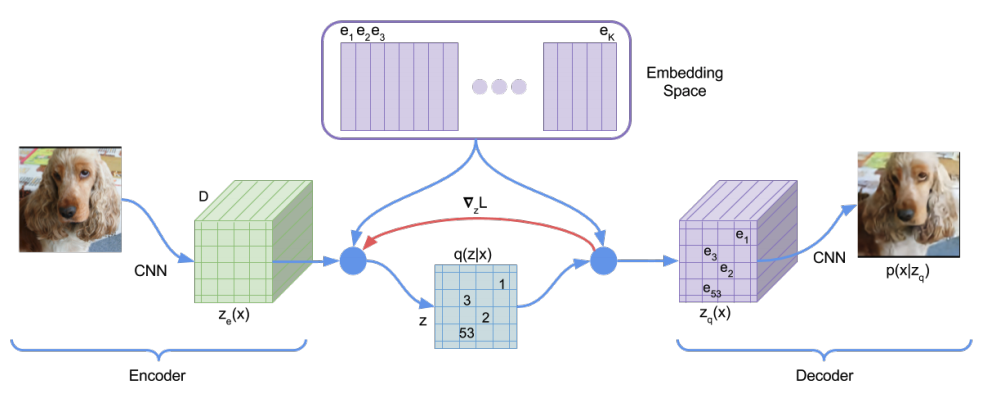}
    \caption{Architecture of VQ-VAE~\cite{van2017neural}. Compared to VAE, the orginal hidden distribution is replaced with a quantized vector dictionary. In addition, the prior distribution is replaced with a pre-trained PixelCNN that models the hierarchical features of images. Taken from~\cite{van2017neural}}
    \label{fig:Cluster}
    \vspace{-0.2cm}
\end{figure}

Recently, a novel and powerful variational AE model called VQ-VAE~\cite{van2017neural} was proposed. VQ-VAE aims to learn discrete latent variables motivated by the fact that many modalities are inherently discrete, such as language, speech, and images. VQ-VAE relies on vector quantization (VQ) to learn the posterior distribution of discrete latent variables. The discrete latent variables are calculated by the nearest neighbor lookup using a shared, learnable embedding table. In training, the gradients are approximated through straight-through estimator~\cite{bengio2013estimating} as

\beq{
\mathcal{L}(x, D(e))=\|x-D(e)\|_2^2+\|sg[E(x)]-e\|^2_2+\beta\|sg[e]-E(x)\|^2_2
}

\noindent where $e$ refers to the codebook, the operator $sg$ refers to a stop-gradient operation that blocks gradients from flowing into its argument, and $\beta$ is a hyperparameter which controls the reluctance to change the code corresponding to the encoder output.

More recently, researchers propose VQ-VAE-2~\cite{razavi2019generating}, which can generate versatile high-fidelity images that rival BigGAN~\cite{brock2018large} on ImageNet~\cite{deng2009imagenet}, the state-of-the-art GAN model. First, the authors enlarge the scale and enhance the autoregressive priors by a powerful PixelCNN~\cite{van2016conditional} prior. Additionally, they adopt a multi-scale hierarchical organization of VQ-VAE, which enables learning local information and global information of images separately. Nowadays, VAE and its variants have been widely used in the computer vision area, such as image representation learning, image generation, video generation.

Variational auto-encoding models have also been employed in node representation learning on graphs. For example, Variational graph auto-encoder (VGAE)~\cite{kipf2016variational} uses the same variational inference technique as VAE with graph convolutional networks (GCN)~\cite{kipf2016semi} as the encoder. Due to the uniqueness of graph-structured data, the objective of VGAE is to reconstruct the adjacency matrix of the graph by measuring node proximity. 
Zhu et al.~\cite{zhu2018deep} propose DVNE, a deep variational network embedding model in Wasserstein space. It learns Gaussian node embedding to model the uncertainty of nodes. 2-Wasserstein distance is used to measure the similarity between the distributions for its effectiveness in preserving network transitivity.
vGraph~\cite{sun2019vgraph} can perform node representation learning and community detection collaboratively through a generative variational inference framework. It assumes that each node can be generated from a mixture of communities, and each community is defined as a multinomial distribution over nodes.

\subsection{Hybrid Generative Models}

\subsubsection{Combining AR and AE Model.}
Some researchers propose to combine the advantages of both AR and AE. MADE~\cite{mathieu2015masked} makes a simple modification to autoencoder. It masks the autoencoder's parameters to respect auto-regressive constraints. Specifically, for the original autoencoder, neurons between two adjacent layers are fully-connected through MLPs. However, in MADE, some connections between adjacent layers are masked to ensure that each input dimension is reconstructed solely from its dimensions. MADE can be easily parallelized on conditional computations, and it can get direct and cheap estimates of high-dimensional joint probabilities by combining AE and AR models.

In NLP, Permutation Language Model (PLM)~\cite{yang2019xlnet} is a representative model that combines the advantage of auto-regressive model and auto-encoding model. XLNet~\cite{yang2019xlnet}, which introduces PLM, is a generalized auto-regressive pretraining method. XLNet enables learning bidirectional contexts by maximizing the expected likelihood over all permutations of the factorization order. To formalize the idea, let $\mathcal{Z_T}$ denotes the set of all possible permutations of the length-$T$ index sequence $[1, 2, ..., T]$, the objective of PLM can be expressed as follows:
\beq{
\max\limits_{\theta}  \mathbb{E}_{\mathbf{z}\sim \mathcal{Z_T}} [ \sum_{t=1}^{T} \log p_{\theta}(x_{z_t} | \mathbf{x}_{\mathbf{z}_{<t}}) ]
}

Actually, for each text sequence, different factorization orders are sampled. Therefore, each token can see its contextual information from both sides. Based on the permuted order, XLNet also conducts reparameterization with positions to let the model know which position is needed to predict. Then a special two-stream self-attention is introduced for target-aware prediction.

\begin{figure}
    \centering
    \includegraphics[width=.48\textwidth]{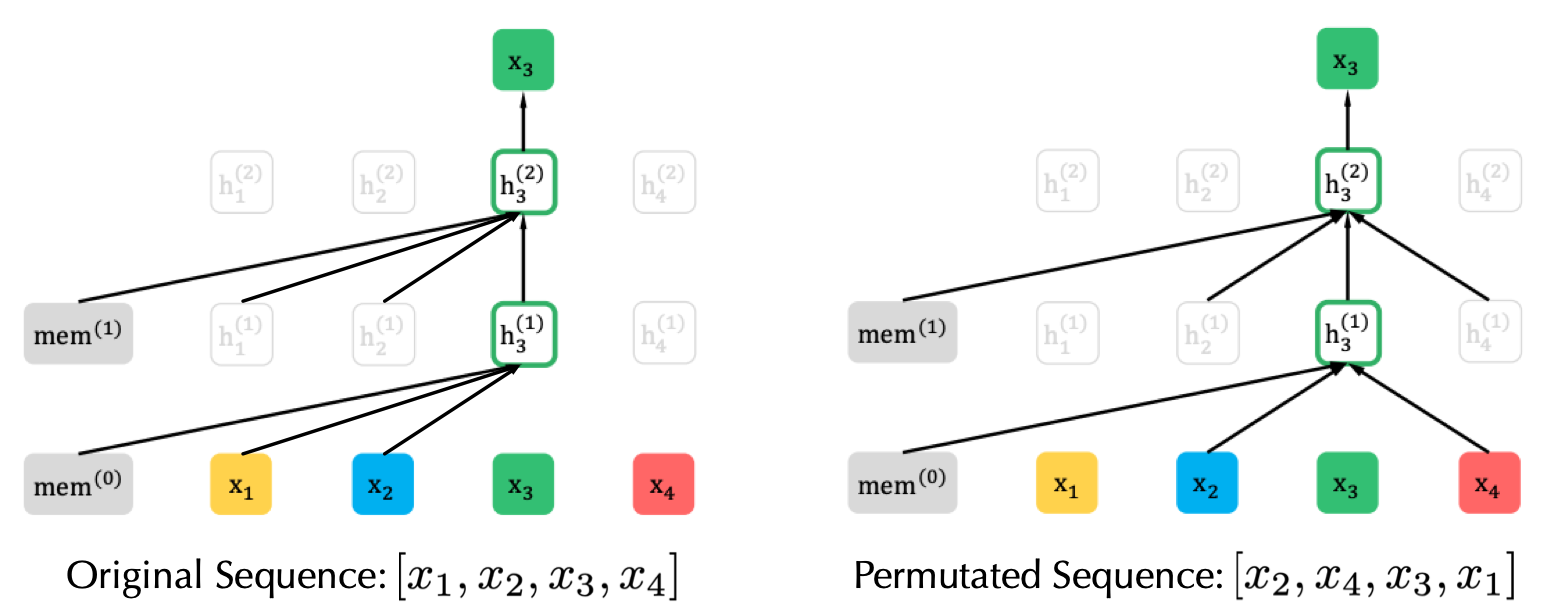}
    \caption{Illustration for permutation language modeling~\cite{yang2019xlnet} objective for predicting $x_3$ given the same input sequence $x$ but with different factorization orders. Adapted from~\cite{yang2019xlnet}}
    \label{fig:xlnet}
    \vspace{-0.2cm}
\end{figure}

Furthermore, different from BERT, inspired by the latest advancements in the AR model, XLNet integrates the segment recurrence mechanism and relative encoding scheme of Transformer-XL~\cite{dai2019transformer} into pre-training, which can model long-range dependency better than Transformer~\cite{vaswani2017attention}.

\subsubsection{Combining AE and Flow-based Models}

In the graph domain, GraphAF~\cite{shi2020graphaf} is a flow-based auto-regressive model for molecule graph generation. It can generate molecules in an iterative process and also calculate the exact likelihood in parallel. GraphAF formalizes molecule generation as a sequential decision process. It incorporates detailed domain knowledge into the reward design, such as valency check. Inspired by the recent progress of flow-based models, it defines an invertible transformation from a base distribution (e.g., multivariate Gaussian) to a molecular graph structure. Additionally, Dequantization technique~\cite{ho2019flow++} is utilized to convert discrete data (including node types and edge types) into continuous data.

\subsection{Pros and Cons} \label{sec:generative_pros_and_cons}
A reason for the generative self-supervised learning's success in self-supervised learning is its ability to recover the original data distribution without assumptions for downstream tasks, which enables generative models' wide applications in both classification and generation. Notably, all the existing generation tasks (including text, image, and audio) rely heavily on generative self-supervised learning. Nevertheless, two shortcomings restrict its performance.

First, despite its central status in generation tasks, generative self-supervised learning is recently found far less competitive than contrastive self-supervised learning in some classification scenarios because contrastive learning's goal naturally conforms the classification objective. Works including MoCo~\cite{he2019momentum}, SimCLR~\cite{chen2020simple}, BYOL~\cite{grill2020bootstrap} and SwAV~\cite{caron2020unsupervised} have presented overwhelming performances on various CV benchmarks. Nevertheless, in the NLP domain, researchers still depend on generative language models to conduct text classification.

Second, the point-wise nature of the generative objective has some inherent defects. This objective is usually formulated as a maximum likelihood function $\mathcal{L}_{MLE}=-\sum_x\mathop{\log} p(x|c)$ where $x$ is all the samples we hope to model, and $c$ is a conditional constraint such as context information. Considering its form, MLE has two fatal problems:

\begin{enumerate}
    \item \textbf{Sensitive and Conservative Distribution}. When $p(x|c)\to0$, $\mathcal{L}_{MLE}$ becomes super large, making generative model extremely sensitive to rare samples. It directly leads to a conservative distribution, which has a low performance.
    \item \textbf{Low-level Abstraction Objective}. In MLE, the representation distribution is modeled at $x$'s level (i.e., point-wise level), such as pixels in images, words in texts, and nodes in graphs. However, most of the classification tasks target at \textit{high-level abstraction}, such as object detection, long paragraph understanding, and molecule classification. 
\end{enumerate}

\noindent and as an opposite approach, generative-contrastive self-supervised learning abandons the point-wise objective. It turns to distributional matching objectives that are more robust and better handle the high-level abstraction challenge in the data manifold.

%% file: 4-contrastive.tex
\section{Contrastive Self-supervised Learning} \label{section:contrastive}
From a statistical perspective, machine learning models are categorized into generative and discriminative models. Given the joint distribution $P(X, Y)$ of the input $X$ and target $Y$, the generative model calculates the $p(X|Y=y)$ while the discriminative model tries to model the $P(Y|X=x)$. Because most of the representation learning tasks hope to model relationships between $x$, for a long time, people believe that the generative model is the only choice for representation learning. 

\begin{figure}[h!]
    \centering
    \includegraphics[width=.48\textwidth]{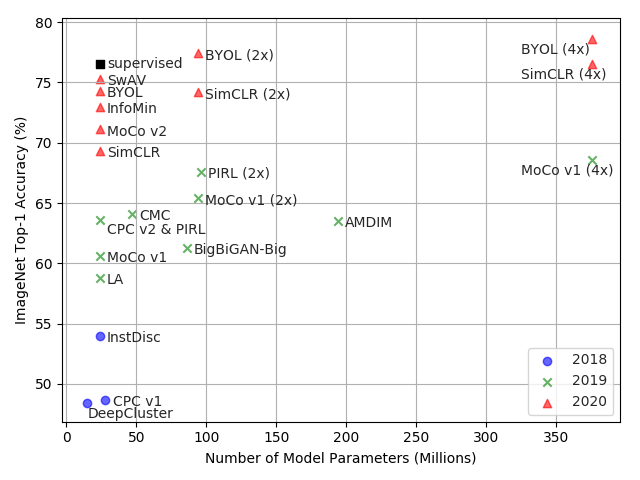}
    \caption{Self-supervised representation learning performance on ImageNet top-1 accuracy in March, 2021, under linear classification protocol. The self-supervised learning's ability on feature extraction is rapidly approaching the supervised method (ResNet50). Except for BigBiGAN, all the models above are contrastive self-supervised learning methods.}
    \label{fig:ImageNet}
    \vspace{-0.2cm}
\end{figure}

However, recent breakthroughs in contrastive learning, such as Deep InfoMax, MoCo and SimCLR, shed light on the potential of discriminative models for representation. Contrastive learning aims at "learn to compare" through a Noise Contrastive Estimation (NCE)~\cite{gutmann2010noise} objective formatted as:

\beq{
\mathcal{L}=\mathbb{E}_{x,x^+,x^-}[-{\rm log}(\frac{e^{f(x)^Tf(x^+)}}{e^{f(x)^Tf(x^+)}+e^{f(x)^Tf(x^-)}}]
}

\noindent where $x^+$ is similar to $x$, $x^-$ is dissimilar to $x$ and $f$ is an encoder (representation function). The similarity measure and encoder may vary from task to task, but the framework remains the same. With more dissimilar pairs involved, we have the InfoNCE~\cite{oord2018representation} formulated as:

\beq{
\mathcal{L}=\mathbb{E}_{x,x^+,x^k}[-{\rm log}(\frac{e^{f(x)^Tf(x^+)}}{e^{f(x)^Tf(x^+)}+\sum_{k=1}^{K}e^{f(x)^Tf(x^k)}}]
}

Here we divide recent contrastive learning frameworks into 2 types: \textit{context-instance} contrast and \textit{instance-instance} contrast. Both of them achieve amazing performance in downstream tasks, especially on classification problems under the linear protocol.

\subsection{\textit{Context-Instance} Contrast}
The context-instance contrast, or so-called \textit{global-local} contrast, focuses on modeling the belonging relationship between the local feature of a sample and its global context representation. When we learn the representation for a local feature, we hope it is associative to the representation of the global content, such as stripes to tigers, sentences to its paragraph, and nodes to their neighborhoods.

There are two main types of Context-Instance Contrast: Predict Relative Position (PRP) and Maximize Mutual Information (MI). The differences between them are:

\begin{itemize}
    \item PRP focuses on learning relative positions between local components. The global context serves as an implicit requirement for predicting these relations (such as understanding what an elephant looks like is critical for predicting relative position between its head and tail).
    \item MI focuses on learning the direct belonging relationships between local parts and global context. The relative positions between local parts are ignored.
\end{itemize}

\subsubsection{Predict Relative Position} \label{spatial}
\begin{figure}[h!]
    \centering
    \includegraphics[width=.48\textwidth]{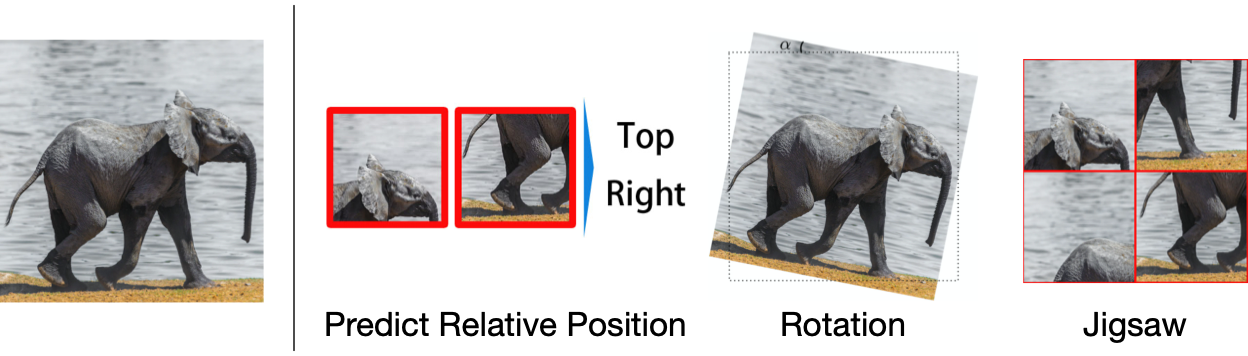}
    \caption{Three typical methods for spatial relation contrast: predict relative position~\cite{doersch2015unsupervised}, rotation~\cite{gidaris2018unsupervised} and solve jigsaw~\cite{noroozi2016unsupervised,wei2019iterative,kim2018learning,misra2019self}.} 
    \label{fig:spatial}
    \vspace{-0.2cm}
\end{figure}

Many data contain rich spatial or sequential relations between parts of it. For example, in image data such as Fig. \ref{fig:spatial}, the elephant's head is on the \textit{right} of its tail. In text data, a sentence like "Nice to meet you." would probably be ahead of "Nice to meet you, too.". Various models regard recognizing relative positions between parts of it as the pretext task~\cite{jing2019self}. It could be to predict relative positions of two patches from a sample~\cite{doersch2015unsupervised}, or to recover positions of shuffled segments of an image (solve jigsaw)~\cite{noroozi2016unsupervised,wei2019iterative,kim2018learning}, or to infer the rotation angle's degree of an image~\cite{gidaris2018unsupervised}. PRP may also serve as tools to create hard positive samples. For instance, the jigsaw technique is applied in PIRL~\cite{misra2019self} to augment the positive sample, but PIRL does not regard solving jigsaw and recovering spatial relation as its objective.

In the pre-trained language model, similar ideas such as Next Sentence Prediction (NSP) are also adopted. NSP loss is initially introduced by BERT~\cite{devlin2019bert}, where for a sentence, the model is asked to distinguish the following and a randomly sampled one. However, some later work empirically proves that NSP helps little, even harm the performance. So in RoBERTa~\cite{liu2019roberta}, the NSP loss is removed.

To replace NSP, ALBERT~\cite{lan2019albert} proposes Sentence Order Prediction (SOP) task. That is because, in NSP, the negative next sentence is sampled from other passages that may have different topics from the current one, turning the NSP into a far easier topic model problem. In SOP, two sentences that exchange their position are regarded as a negative sample, making the model concentrate on the semantic meaning's coherence.

\hide{Though some good results are observed, it is still doubtful that how much adopting spatial relation as pretext task could contribute to downstream tasks because distinguishable features usually matter more in downstream classification tasks rather than relative positions. However, it is expected to see if this kind of model could help in fields that heavily depend on spatial relations, such as visual question answering.}

\subsubsection{Maximize Mutual Information}
This kind of method derives from mutual information (MI) -- a fundamental concept in statistics.  Mutual information targets modeling the association between two variables, and our objective is to maximize it. Generally, this kind of models optimize

\beq{
    \max\limits_{g_1\in \mathcal{G}_1,g_2\in \mathcal{G}_1} I(g_1(x_1), g_2(x_2))
}

\noindent where $g_i$ is the representation encoder, $\mathcal{G}_i$ is a class of encoders with some constraints, and $I(\cdot,\cdot)$ is a sample-based estimator for the accurate mutual information. In applications, MI is notorious for its complex computation. A common practice is to alternatively maximize $I$'s lower bound with an NCE objective.

\begin{figure}[h!]
    \centering
    \includegraphics[width=.48\textwidth]{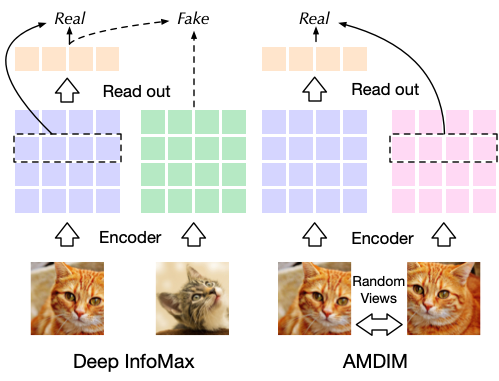}
    \caption{Two representatives for mutual information's application in contrastive learning. Deep InfoMax (DIM)~\cite{hjelm2018learning} first encodes an image into feature maps, and leverage a read-out function (or so-called summary function) to produce a summary vector. AMDIM~\cite{bachman2019learning} enhances the DIM through randomly choosing another view of the image to produce the summary vector.}
    \label{fig:DIM}
    \vspace{-0.2cm}
\end{figure}

Deep InfoMax~\cite{hjelm2018learning} is the first one to explicitly model mutual information through a contrastive learning task, which maximize the MI between a local patch and its global context. For real practices, take image classification as an example, we can encode a cat image $x$ into $f(x)\in\mathbb{R}^{M\times M\times d}$, and take out a local feature vector $v\in\mathbb{R}^{d}$. To conduct contrast between instance and context, we need two other things:

\begin{itemize}
    \item a summary function $g{\rm :} \mathbb{R}^{M\times M\times d}\to\mathbb{R}^{d}$ to generate the context vector $s=g(f(x))\in\mathbb{R}^{d}$
    \item another cat image $x^-$ and its context vector $s^-=g(f(x^-))$.
\end{itemize}

\noindent and the contrastive objective is then formulated as

\beq{
\mathcal{L} = \mathbb{E}_{v, x}[-{\rm log}(\frac{e^{v^T\cdot s}}{e^{v^T\cdot s}+e^{v^T\cdot s^-}})]
}

Deep InfoMax provides us with a new paradigm and boosts the development of self-supervised learning. The first influential follower is Contrastive Predictive Coding (CPC)~\cite{oord2018representation} for speech recognition. CPC maximizes the association between a segment of audio and its context audio. To improve data efficiency, it takes several negative context vectors at the same time. Later on, CPC has also been applied in image classification.

AMDIM~\cite{bachman2019learning} enhances the positive association between a local feature and its context. It randomly samples two different views of an image (truncated, discolored, and so forth) to generate the local feature vector and context vector, respectively. CMC~\cite{tian2019contrastive} extends it into several different views for one image and samples another irrelevant image as the negative. However, CMC is fundamentally different from Deep InfoMax and AMDIM because it proposes to measure the instance-instance similarity rather than context-instance similarity. We will discuss it in the following subsection.

In language pre-training, InfoWord~\cite{kong2019mutual} proposes to maximize the mutual information between a global representation of a sentence and \textit{n-grams} in it. The context is induced from the sentence with selected n-grams being masked, and the negative contexts are randomly picked out from the corpus.
\begin{figure}[h!]
    \centering
    \includegraphics[width=.48\textwidth]{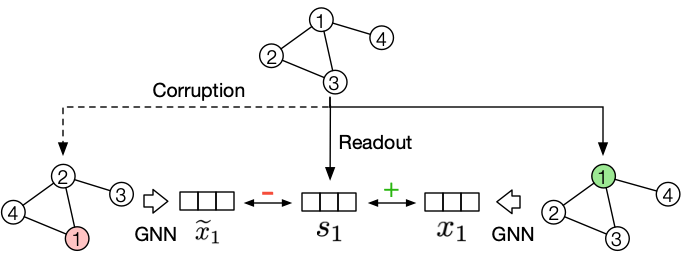}
    \caption{Deep Graph InfoMax~\cite{velivckovic2018deep} uses a readout function to generate summary vector $s_1$, and puts it into a discriminator with node 1's embedding $x_1$ and corrupted embedding $\widetilde{x}_1$ respectively to identify which embedding is the real embedding. The corruption is to shuffle the positions of nodes.}
    \label{fig:DGI}
    \vspace{-0.2cm}
\end{figure}

In graph learning, Deep Graph InfoMax (DGI)~\cite{velivckovic2018deep} regards a node's representation as the local feature and the average of randomly samples 2-hop neighbors as the context. However, it is hard to generate negative contexts on a single graph. To solve this problem, DGI proposes to \textit{corrupt} the original context by keeping the sub-graph structure and permuting the node features. DGI is followed by many works, such as InfoGraph~\cite{sun2019infograph}, which targets learning graph-level representation rather than node level, maximizing the mutual information between graph-level representation and substructures at different levels. As what CMC has done to improve Deep InfoMax, in~\cite{hassani2020contrastive} authors propose a contrastive multi-view representation learning method for the graph. They also discover that graph diffusion is the most effective way to yield augmented positive sample pairs in graph learning.

As an attempt to unify graph pre-training, in~\cite{hu2019strategies}, the authors systematically analysis the pre-training strategies for graph neural networks from two dimensions: attribute/structural and node-level/graph-level. For structural prediction at node-level, they propose Context Prediction to maximize the MI between the k-hop neighborhood's representations and its context graph. For attributes in the chemical domain, they propose Attribute Mask to predict a node's attribute according to its neighborhood, which is a generative objective similar to token masks in BERT.

S$^{\rm 2}$GRL~\cite{peng2020self} further separates nodes in the context graph into k-hop context subgraphs and maximizes their MI with target node, respectively. However, a fundamental problem of graph pre-training is about learning inductive biases across graphs, and existing graph pre-training work is only applicable for a specific domain.

\subsection{\textit{Instance-Instance} Contrast}
Though MI-based contrastive learning achieves great success, some recent studies~\cite{tschannen2019mutual,he2019momentum,chen2020improved,chen2020simple} cast doubt on the actual improvement brought by MI. 

The~\cite{tschannen2019mutual} provides empirical evidence that the success of the models mentioned above is only loosely connected to MI by showing that an upper bound MI estimator leads to ill-conditioned and lower performance representations. Instead, more should be attributed to encoder architecture and a negative sampling strategy related to metric learning. A significant focus in metric learning is to perform hard positive sampling while increasing the negative sampling efficiency. They probably play a more critical role in MI-based models' success.

As an alternative, instance-instance contrastive learning discards MI and directly studies the relationships between different samples' instance-level local representations as what metric learning does. Instance-level representation, rather than context-level, is more crucial for a wide range of classification tasks. For example, in an image classified as ``dog'', while there must be dog instances, some other irrelevant context objects such as grass might appear. But what matters for the image classification is the dog rather than the context. Another example would be sentence emotional classification, which primarily relies on few but important keywords.

In the early stage of instance-instance contrastive learning's development, researchers borrow ideas from semi-supervised learning to produce pseudo labels via cluster-based discrimination and achieve rather good performance on representations. More recently, CMC~\cite{tian2019contrastive}, MoCo~\cite{he2019momentum}, SimCLR~\cite{chen2020simple}, and BYOL~\cite{grill2020bootstrap} further support the above conclusion by outperforming the context-instance contrastive methods and achieve a competitive result to supervised methods under the linear classification protocol. We will start with cluster-based discrimination proposed earlier and then turn to instance-based discrimination.

\subsubsection{Cluster Discrimination} \label{sec:cluster-based}
 
Instance-instance contrast is first studied in clustering-based methods~\cite{yang2016joint,li2016unsupervised,noroozi2018boosting,caron2018deep}, especially the DeepCluster~\cite{caron2018deep} which first achieves competitive performance to the supervised model AlexNet~\cite{krizhevsky2012imagenet}. 

\begin{figure}[t!]
    \centering
    \includegraphics[width=.43\textwidth]{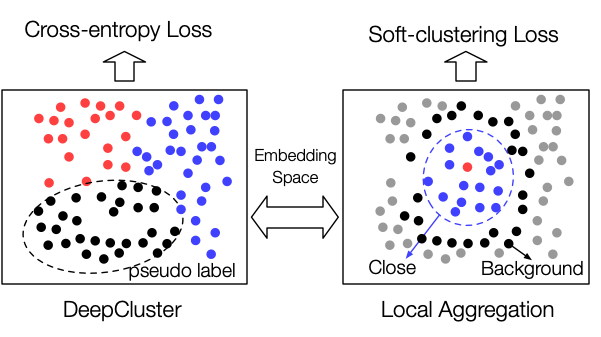}
    \caption{Cluster-based instance-instance contrastive emthods: DeepCluster~\cite{caron2018deep} and Local Aggregation~\cite{zhuang2019local}. In the embedding space, DeepCluster uses clustering to yield pseudo labels for discrimination to draw near similar samples. However, Local Aggregation shows that a egocentric soft-clustering objective would be more effective.}
    \label{fig:Cluster}
    \vspace{-0.2cm}
\end{figure}

Image classification asks the model to categorize images correctly, and the representation of images in the same category should be similar. Therefore, the motivation is to pull similar images near in the embedding space. In supervised learning, this pulling-near process is accomplished via label supervision; in self-supervised learning, however, we do not have such labels. To solve the label problem, Deep Cluster~\cite{caron2018deep} proposes to leverage clustering to yield pseudo labels and asks a discriminator to predict images' labels. The training could be formulated in two steps. In the first step, DeepCluster uses K-means to cluster encoded representation and produces pseudo labels for each sample. Then in the second step, the discriminator predicts whether two samples are from the same cluster and back-propagates to the encoder. These two steps are performed iteratively.

Recently, Local Aggregation (LA)~\cite{zhuang2019local} has pushed forward the cluster-based method's boundary. It points out several drawbacks of DeepCluster and makes the corresponding optimization. First, in DeepCluster, samples are assigned to mutual-exclusive clusters, but LA identifies neighbors separately for each example. Second, DeepCluster optimizes a cross-entropy discriminative loss, while LA employs an objective function that directly optimizes a local soft-clustering metric. These two changes substantially boost the performance of LA representation on downstream tasks.

A similar work to LA would be VQ-VAE~\cite{van2017neural,razavi2019generating} that we introduce in Section \ref{section:generative}. To conquer the traditional deficiency for VAE to generate high-fidelity images, VQ-VAE proposes quantizing vectors. For the feature matrix encoded from an image, VQ-VAE substitutes each 1-dimensional vector in the matrix to the nearest one in an embedding dictionary. This process is somehow the same as what LA is doing.

Clustering-based discrimination may also help in the generalization of other pre-trained models, transferring models from pretext objectives to downstream tasks better. Traditional representation learning models have only two stages: one for pre-training and the other for evaluation. ClusterFit~\cite{yan2019clusterfit} introduces a cluster prediction fine-tuning stage similar to DeepCluster between the above two stages, which improves the representation's performance on downstream classification evaluation. 

Despite the previous success of cluster discrimination-based contrastive learning, the two-stage training paradigm is time-consuming and poor performing compared to later instance discrimination-based methods, including CMC~\cite{tian2019contrastive}, MoCo~\cite{he2019momentum} and SimCLR~\cite{chen2020simple}. These instance discrimination-based methods have got rid of the slow clustering stage and introduced efficient data augmentation (i.e., multi-view) strategies to boost the performance. In light of these problems, authors in SwAV~\cite{caron2020unsupervised} bring online clustering ideas and multi-view data augmentation strategies into the cluster discrimination approach. SwAV proposes a swapped prediction contrastive objectives to deal with multi-view augmentation. The intuition is that, given some (clustered) prototypes, different views of the same images should be assigned into the same prototypes. SwAV names this ``assignment'' as ``codes''. To accelerate code computing, the authors of SwAV design an online computing strategy. SwAV outperforms instance discrimination-based methods when model size is small and is more computationally efficient. Based on SwAV, a 1.3-billion-parameter SEER~\cite{goyal2021selfsupervised} is trained on 1 billion web images collected from Instagram.

In graph learning, M3S~\cite{sun2019multi} adopts a similar idea to perform DeepCluster-style self-supervised pre-training for better semi-supervised prediction. Given little labeled data and many unlabeled data, for every stage, M3S first pre-train itself to produce pseudo labels on unlabeled data as DeepCluster does and then compares these pseudo labels with those predicted by the model being supervised trained on labeled data. Only top-k confident labels are added into a labeled set for the next stage of semi-supervised training. In~\cite{you2020does}, this idea is further developed into three pre-training tasks: topology partitioning (similar to spectral clustering), node feature clustering, and graph completion.

\hide{
\begin{figure}
    \centering
    \includegraphics[width=.48\textwidth]{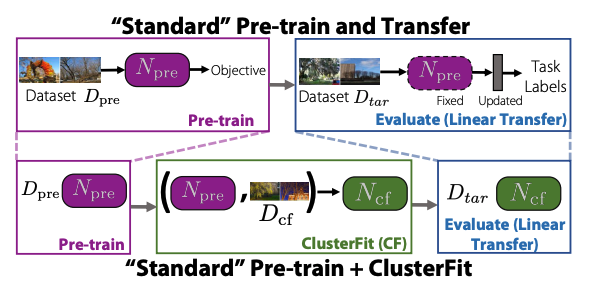}
    \caption{Architecture of CluterFit~\cite{yan2019clusterfit}. A second stage of cluster discrimination pre-training is inserted after the original pre-training stage.}
    \label{fig:ClusterFit}
    \vspace{-0.2cm}
\end{figure}
}

\subsubsection{Instance Discrimination}
The prototype of leveraging instance discrimination as a pretext task is InstDisc~\cite{wu2018unsupervised}. Based on InstDisc, CMC~\cite{tian2019contrastive} proposes to adopt multiple different views of an image as positive samples and take another one as the negative. CMC draws near multiple views of an image in the embedding space and pulls away from other samples. However, it is somehow constrained by the idea of Deep InfoMax, which only samples one negative sample for each positive one.

In MoCo~\cite{he2019momentum}, researchers further develop the idea of leveraging \textit{instance discrimination} via momentum contrast, which substantially increases the amount of negative samples. For example, given an input image $x$, our intuition is to learn a instinct representation $q = f_q(x)$ by a query encoder $f_q(\cdot)$ that can distinguish $x$ from any other images. Therefore, for a set of other images $x_i$, we employ an asynchronously updated key encoder $f_k(\cdot)$ to yield $k_+=f_k(x)$ and $k_i=f_k(x_i)$, and optimize the following objective

\beq{
    \mathcal{L}=-{\rm log}\frac{{\rm exp}(q\cdot k_+/\tau)}{\sum^K_{i=0}{\rm exp}(q\cdot k_i/\tau)}
}

\noindent where $K$ is the number of negative samples. This formula is in the form of InfoNCE.

Besides, MoCo presents two other critical ideas in dealing with negative sampling efficiency.

\begin{itemize}
    \item First, it abandons the traditional end-to-end training framework. It designs the momentum contrast learning with two encoders (query and key), which prevents the fluctuation of loss convergence in the beginning period. 
    \item Second, to enlarge negative samples' capacity, MoCo employs a queue (with K as large as 65536) to save the recently encoded batches as negative samples. This significantly improves the negative sampling efficiency.
\end{itemize}

\begin{figure}[h!]
    \centering
    \includegraphics[width=.48\textwidth]{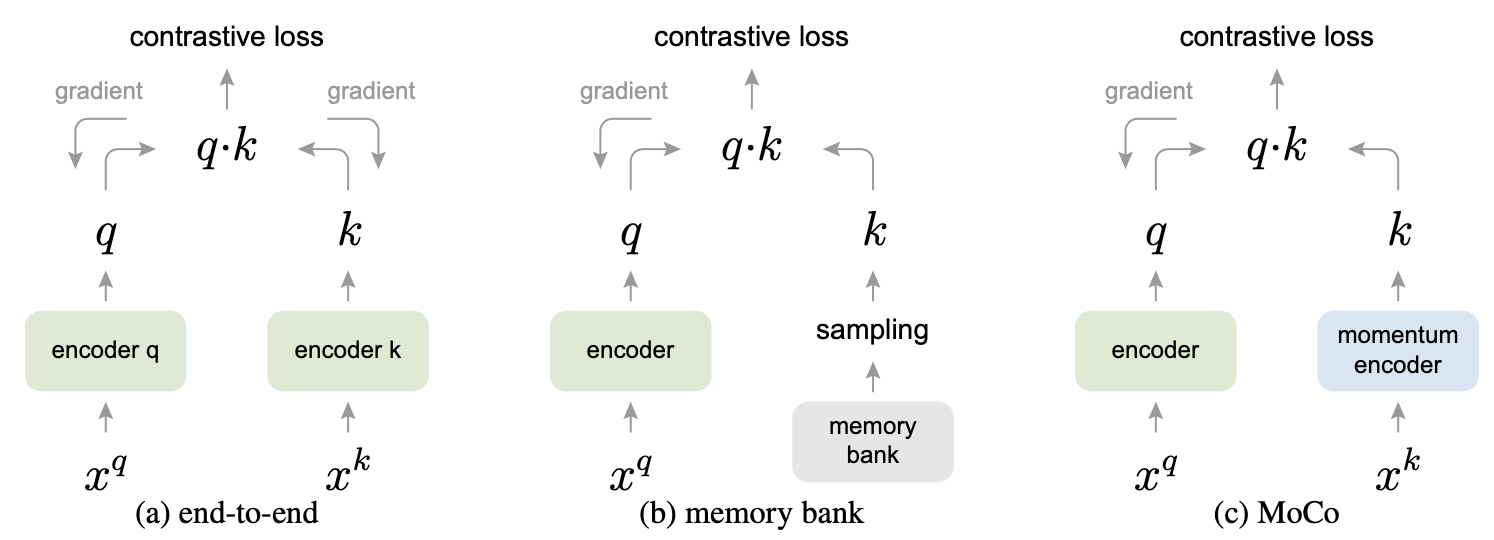}
    \caption{Conceptual comparison of three contrastive loss mechanisms. Taken from MoCo~\cite{he2019momentum}.}
    \label{fig:MoCo}
    \vspace{-0.2cm}
\end{figure}

There are some other auxiliary techniques to ensure the training convergence, such as batch shuffling to avoid trivial solutions and temperature hyper-parameter $\tau$ to adjust the scale.

However, MoCo adopts a too simple positive sample strategy: a pair of positive representations come from the same sample without any transformation or augmentation, making the positive pair far too easy to distinguish. PIRL~\cite{misra2019self} adds jigsaw augmentation as described in Section \ref{spatial}. PIRL asks the encoder to regard an image and its jigsawed one as similar pairs to produce a pretext-invariant representation. 

\begin{figure}[h!]
    \centering
    \includegraphics[width=.48\textwidth]{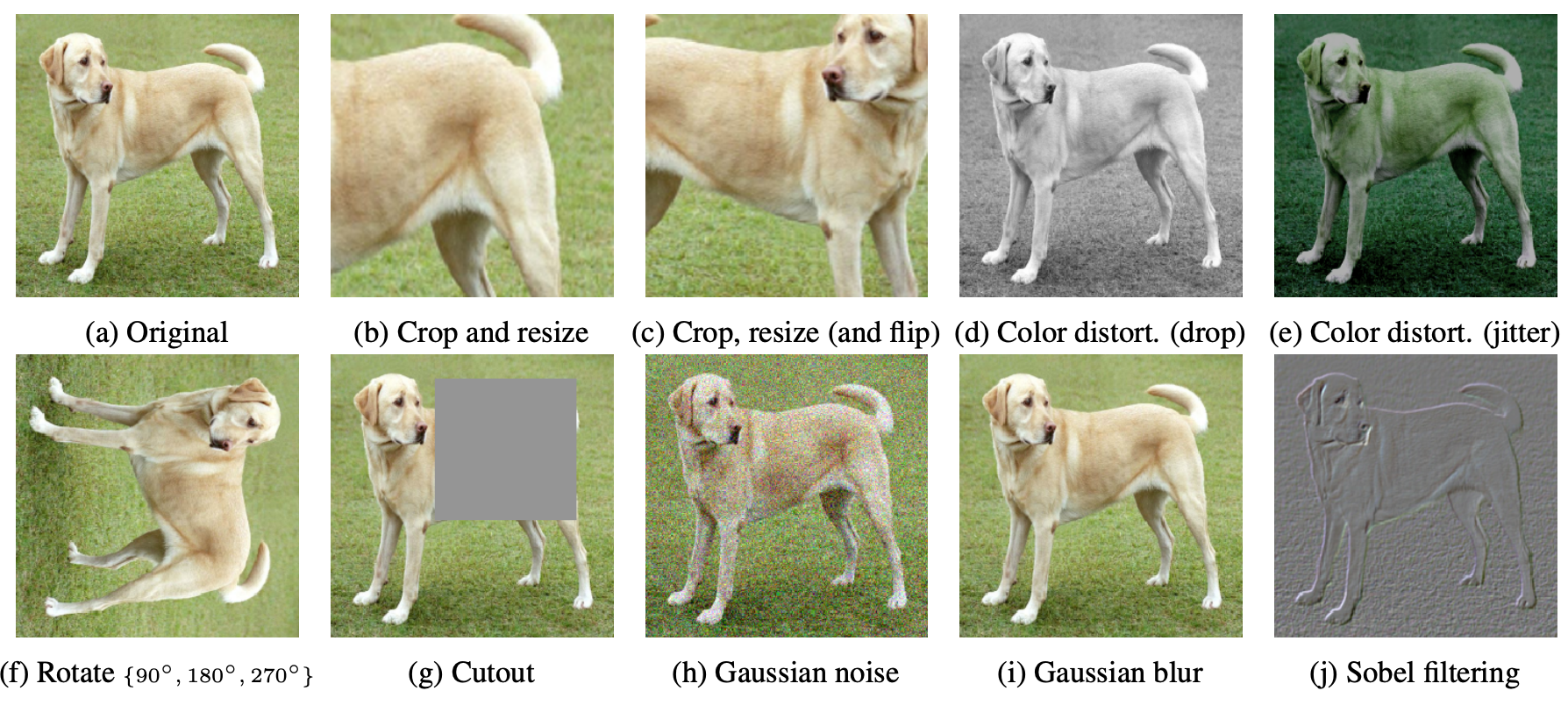}
    \caption{Ten different views adopted by SIMCLR~\cite{chen2020simple}. The enhancement of positive samples substantially improves the self-supervised learning performance. Taken from~\cite{chen2020simple}}
    \label{fig:SIMCLR}
    \vspace{-0.2cm}
\end{figure}

In SimCLR~\cite{chen2020simple}, the authors further illustrate the importance of a hard positive sample strategy by introducing data augmentation in 10 forms. This data augmentation is similar to CMC~\cite{tian2019contrastive}, which leverages several different views to augment the positive pairs. SimCLR follows the end-to-end training framework instead of momentum contrast from MoCo, and to handle the large-scale negative samples problem, SimCLR chooses a batch size of $N$ as large as 8196.

The details are as follows. A minibatch of $N$ samples is augmented to be $2N$ samples $\hat{x}_j(j=1,2,...,2N)$. For a pair of a positive sample $\hat{x}_i$ and $\hat{x}_j$ (derive from one original sample), other $2(N-1)$ are treated as negative ones. A pairwise contrastive loss NT-Xent loss~\cite{chen2017sampling} is defined as

\beq{
l_{i,j}=-{\rm log}\frac{{\rm exp}({\rm sim}(\hat{x}_i,\hat{x}_j)/\tau)}{\sum_{k=1}^{2N}\mathbb{I}_{[k\neq i]}{\rm exp}({\rm sim}(\hat{x}_i,\hat{x}_k)/\tau)}
}

\noindent noted that $l_{i,j}$ is asymmetrical, and the ${\rm sim}(\cdot,\cdot)$ function here is a cosine similarity function that can normalize the representations. The summed up loss is 

\beq{
\mathcal{L}=\frac{1}{2N}\sum_{k=1}^N[l_{2i-1,2i}+l_{2i,2i-1}]
}

SimCLR also provides some other practical techniques, including a learnable nonlinear transformation between the representation and the contrastive loss, more training steps, and deeper neural networks. \cite{chen2020improved} conducts ablation studies to show that techniques in SimCLR can also further improve MoCo's performance.

More investigation into augmenting positive samples is made in InfoMin~\cite{tian2020makes}. The authors claim that we should select those views with less mutual information for better-augmented views in contrastive learning. In the optimal situation, the views should only share the label information. To produce such optimal views, the authors first propose an unsupervised method to minimize mutual information between views. However, this may result in a loss of information for predicting labels (such as a pure blank view). Therefore, a semi-supervised method is then proposed to find views sharing only label information. This technique leads to an improve about 2\% over MoCo v2.

A more radical step is made by BYOL~\cite{grill2020bootstrap}, which discards negative sampling in self-supervised learning but achieves an even better result over InfoMin. For contrastive learning methods we mentioned above, they learn representations by predicting different views of the same image and cast the prediction problem directly in representation space. However, predicting directly in representation space can lead to collapsed representations because multi-views are generally \textit{too predictive} for each other. Without negative samples, it would be too easy for the neural networks to distinguish those positive views.

\begin{figure}[h!]
    \vspace{-0.3cm}
    \centering
    \includegraphics[width=.48\textwidth]{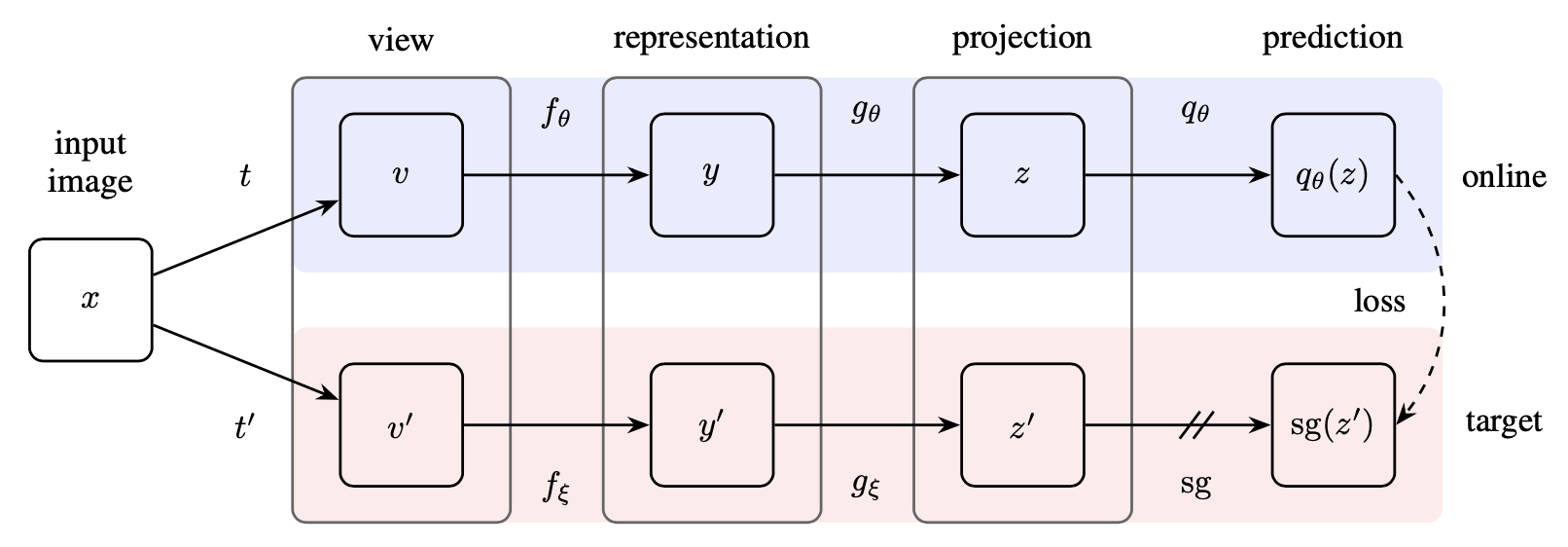}
    \caption{The architecture of BYOL~\cite{grill2020bootstrap}. Noted that the online encoder has an additional layer $q_\theta$ compared to the target one, which gives the representations some flexibility to be improved during the training. Taken from~\cite{grill2020bootstrap}}
    \label{fig:BYOL}
    \vspace{-0.2cm}
\end{figure}

In BYOL, researchers argue that negative samples may not be necessary in this process. They show that, if we use a fixed randomly initialized network (which would not collapse because it is not trained) to serve as the key encoder, the representation produced by query encoder would still be improved during training. If then we set the target encoder to be the trained query encoder and iterate this procedure, we would progressively achieve better performance. Therefore, BYOL proposes an architecture (Figure \ref{fig:BYOL}) with an exponential moving average strategy to update the target encoder just as MoCo does. Additionally, instead of using cross-entropy loss, they follow the regression paradigm in which mean square error is used as:

\beq{
\mathcal{L}_\theta^{\rm BYOL}\triangleq ||\bar{q_\theta}(z_\theta)-\bar{z_\xi}^\prime||_2^2=2-2\cdot\frac{\langle q_\theta(z_\theta),z_\xi^\prime\rangle}{||q_\theta(z_\theta)||_2\cdot ||z_\xi^\prime||_2}
}

This not only makes the model better-performed in downstream tasks, but also more robust to smaller batch size. In MoCo and SimCLR, a drop in batch size results in a significant decline in performance. However, in BYOL, although batch size still matters, it is far less critical. The ablation study shows that a batch size of 512 only causes a drop of 0.3\% compared to a standard batch size of 4096, while SimCLR shows a drop of 1.4\%.

In SimSiam~\cite{chen2020exploring}, researchers further study how necessary is negative sampling, and even batch normalization in contrastive representation learning. They show that the most critical component in BYOL is the \textit{stop gradient} operation, which makes the target representation stable. SimSiam is proved to converge faster than MoCo, SimCLR, and BYOL with even smaller batch sizes, while the performance only slightly decreases.

Some other works are inspired by theoretical analysis into the contrastive objective. ReLIC~\cite{mitrovic2020representation} argues that contrastive pre-training teaches the encoder to causally disentangle the invariant content (i.e., main objects) and style (i.e., environments) in an image. To better enforce this observation in the data augmentation, they propose to add an extra KL-divergence regularizer between prediction logits of an image's different views. The results show that this can enhance the models' generalization ability and robustness and improve the performance.

In graph learning, Graph Contrastive Coding (GCC)~\cite{qiu2020gcc} is a pioneer to leverage instance discrimination as the pretext task for structural information pre-training. For each node, we sample two subgraphs independently by random walks with restart and use top eigenvectors from their normalized graph Laplacian matrices as nodes' initial representations. Then we use GNN to encode them and calculate the InfoNCE loss as what MoCo and SimCLR do, where the node embeddings from the same node (in different subgraphs) are viewed as similar. Results show that GCC learns better transferable structural knowledge than previous work such as struc2vec~\cite{ribeiro2017struc2vec}, GraphWave~\cite{donnat2018learning} and ProNE~\cite{zhang2019prone}. GraphCL~\cite{you2020graph} studies the data augmentation strategies in graph learning. They propose four different augmentation methods based on edge perturbation and node dropping. It further demonstrates that the appropriate combination of these strategies can yield even better performance.

\subsection{Self-supervised Contrastive Pre-training for Semi-supervised Self-training}
While contrastive learning-based self-supervised learning continues to push the boundaries on various benchmarks, labels are still important because there is a gap between training objectives of self-supervised learning and supervised learning. In other words, no matter how self-supervised learning models improve, they are still the only powerful feature extractor, and to transfer to the downstream task, we still need labels more or less. As a result, to bridge the gap between self-supervised pre-training and downstream tasks, semi-supervised learning is what we are looking for.

Recall the MoCo~\cite{he2019momentum} that have topped the ImageNet leader-board. Although it is proved beneficial for many other downstream vision tasks, it fails to improve the COCO object detection task. Some following work~\cite{newell2020useful,zoph2020rethinking} investigates this problem and attributes it to the gap between the instance discrimination and object detection. In such a situation, while pure self-supervised pre-training fails to help, semi-supervised-based self-training can contribute a lot to it.

First, we will clarify the definitions of semi-supervised learning and self-training. Semi-supervised learning is an approach to machine learning that combines a small amount of labeled data with many unlabeled data during training. Various methods derive from several different assumptions made on the data distribution, with self-training (or self-labeling) being the oldest. In self-training, a model is trained on the small amount of labeled data and then yield labels on unlabeled data. Only those data with highly confident labels are combined with original labeled data to train a new model. We iterate this procedure to find the best model.

The current state-of-the-art supervised model~\cite{xie2020self} on ImageNet follows the self-training paradigm, where we first train an EfficientNet model on labeled ImageNet images and use it as a teacher to generate pseudo labels on 300M unlabeled images. We then
train a larger EfficientNet as a student model based on labeled and pseudo labeled images. We iterate this process by putting back the student as the teacher. During the pseudo labels generation, the teacher is not noised so that the pseudo labels are as accurate as possible. However, during the student's learning, we inject noise such as dropout, stochastic depth, and data augmentation via RandAugment to the student to generalize better than the teacher.

In light of semi-supervised self-training's success, it is natural to rethink its relationship with the self-supervised methods, especially with the successful contrastive pre-trained methods. In Section \ref{sec:cluster-based}, we have introduced M3S~\cite{sun2020multi} that attempts to combine cluster-based contrastive pre-training and downstream semi-supervised learning. For computer vision tasks, Zoph et al.~\cite{zoph2020rethinking} study the MoCo pre-training and a self-training method in which a teacher is first trained on a downstream dataset (e.g., COCO) and then yield pseudo labels on unlabeled data (e.g., ImageNet), and finally a student learns jointly over real labels on the downstream dataset and pseudo labels on unlabeled data. They surprisingly find that pre-training's performance hurts while self-training still benefits from strong data augmentation. Besides, more labeled data diminishes the value of pre-training, while semi-supervised self-training always improves. They also discover that the improvements from pre-training and self-training are orthogonal to each other, i.e., contributing to the performance from different perspectives. The model with joint pre-training and self-training is the best.

Chen et al.~\cite{chen2020big}'s SimCLR v2 supports the conclusion mentioned above by showing that with only 10\% of the original ImageNet labels, the ResNet-50 can surpass the supervised one with joint pre-training and self-training. They propose a 3-step framework:

\begin{enumerate}
    \item Do self-supervised pre-training as SimCLR v1, with some minor architecture modification and a deeper ResNet.
    \item Fine-tune the last few layers with only 1\% or 10\% of original ImageNet labels.
    \item Use the fine-tuned network as \textit{teacher} to yield labels on unlabeled data to train a smaller \textit{student} ResNet-50.
\end{enumerate}

The success in combining self-supervised contrastive pre-training and semi-supervised self-training opens up our eyes for a future data-efficient deep learning paradigm. More work is expected for investigating their latent mechanisms.

\subsection{Pros and Cons}
Because contrastive learning has assumed the downstream applications to be classifications, it only employs the encoder and discards the decoder in the architecture compared to generative models. Therefore, contrastive models are usually light-weighted and perform better in discriminative downstream applications. 

Contrastive learning is closely related to metric learning, a discipline that has been long studied. However, self-supervised contrastive learning is still an emerging field, and many problems remain to be solved, including:

\begin{enumerate}
    \item \textbf{Scale to natural language pre-training.} Despite its success in computer vision, contrastive pre-training does not present a convincing result in the NLP benchmarks. Most contrastive learning in NLP now lies in BERT's supervised fine-tuning, such as improving BERT's sentence-level representation~\cite{reimers2019sentence}, information retrieval~\cite{karpukhin2020dense}. Few algorithms have been proposed to apply contrastive learning in the pre-training stage. As most language understanding tasks are classifications, a contrastive language pre-training approach should be better than the current generative language models.
    \item \textbf{Sampling efficiency.} Negative sampling is a must for most contrastive learning, but this process is often tricky, biased, and time-consuming. BYOL~\cite{grill2020bootstrap} and SimSiam~\cite{chen2020exploring} are the pioneers to get contrastive learning rid of negative samples, but it can be improved. It is also not clear enough that what role negative sampling plays in contrastive learning.
    \item \textbf{Data augmentation.} Researchers have proved that data augmentation can boost contrastive learning's performance, but the theory for why and how it helps is still quite ambiguous. This hinders its application into other domains, such as NLP and graph learning, where the data is discrete and abstract.
\end{enumerate}

%% file: 5-generative-contrastive.tex
\section{Generative-Contrastive (Adversarial) Self-supervised Learning} \label{section:generative-contrastive}
Generative-contrastive representation learning, or in a more familiar name \textit{adversarial representation learning}, leverage discriminative loss function as the objective. Yann Lecun comments on adversarial learning as "the most interesting idea in the last ten years in machine learning.". Its application in learning representation is also booming.

The idea of adversarial learning derives from generative learning, where researchers have observed some inherent shortcomings of point-wise generative reconstruction (See Section \ref{sec:generative_pros_and_cons}). As an alternative, adversarial learning learns to reconstruct the original data distribution rather than the samples by minimizing the distributional divergence. 

In terms of contrastive learning, adversarial methods still preserve the generator structure consisting of an encoder and a decoder. In contrast, the contrastive abandons the decoder component (as shown in Fig. \ref{fig:ad_three}). It is critical because, on the one hand, the generator endows adversarial learning with strong expressiveness that is peculiar to generative models; on the other hand, it also makes the objective of adversarial methods far more challenging to learn than that of contrastive methods, leading to unstable convergence. In the adversarial setting, the decoder's existence asks the representation to be "reconstructive," in other words, it contains all the necessary information for constructing the inputs. However, in the contrastive setting, we only need to learn "distinguishable" information to discriminate different samples.

To sum up, the adversarial methods absorb merits from both generative and contrastive methods together with some drawbacks. In a situation where we need to fit on an implicit distribution, it is a better choice. In the following several subsections, we will discuss its various applications on representation learning.

\subsection{Generate with Complete Input}

This section introduces GAN and its variants for representation learning, focusing on capturing the sample's complete information.

The inception of adversarial representation learning should be attributed to Generative Adversarial Networks (GAN)~\cite{radford2015unsupervised}, which proposes the adversarial training framework. Follow GAN, many variants~\cite{ledig2017photo,pathak2016context,iizuka2017globally,brock2018large,karras2019style,isola2017image} emerge and reshape people's understanding of deep learning's potential. GAN's training process could be viewed as two players play a game; one generates fake samples while another tries to distinguish them from real ones. To formulate this problem, we define $G$ as the generator, $D$ as the discriminator, $p_{data}(x)$ as the real sample distribution, $p_z(z)$ as the learned latent sample distribution, we want to optimize this min-max game

\beq{
    \min\limits_G\max\limits_D\mathbb{E}_{x\sim p_{data}(x)}[{\rm log}D(x)]+\mathbb{E}_{z\sim p_{z}(z)}[{\rm log}(1 - D(G(z)))]
}

Before VQ-VAE2, GAN maintains dominating performance on image generation tasks over purely generative models, such as autoregressive PixelCNN and autoencoder VAE. It is natural to think about how this framework could benefit representation learning. 

However, there is a gap between generation and representation. Compared to autoencoder's explicit latent sample distribution $p_z(z)$, GAN's latent distribution $p_z(z)$ is implicitly modeled. We need to extract this implicit distribution out. To bridge this gap, AAE~\cite{makhzani2015adversarial} first proposes a solution to follow the autoencoder's natural idea. The generator in GAN could be viewed as an implicit autoencoder. We can replace the generator with an explicit variational autoencoder (VAE) to extract the representation out. Recall the objective of VAE

\beq{
    \mathcal{L}_{\rm VAE}=-\mathbb{E}_{q(z|x)}(-{\rm log}(p(x|z))+{\rm KL}(q(z|x)\|p(z))
}

\noindent As we mentioned before, compared to $l2$ loss of autoencoder, discriminative loss in GAN better models the high-level abstraction. To alleviate the problem, AAE substitutes the KL divergence function for a discriminative loss 
\beq{\mathcal{L}_{Disc}={\rm CrossEntropy}(q(z),p(z))
}

\noindent that asks the discriminator to distinguish representation from the encoder and a prior distribution.

\hide{
\begin{figure}[t!]
    \centering
    \includegraphics[width=.38\textwidth]{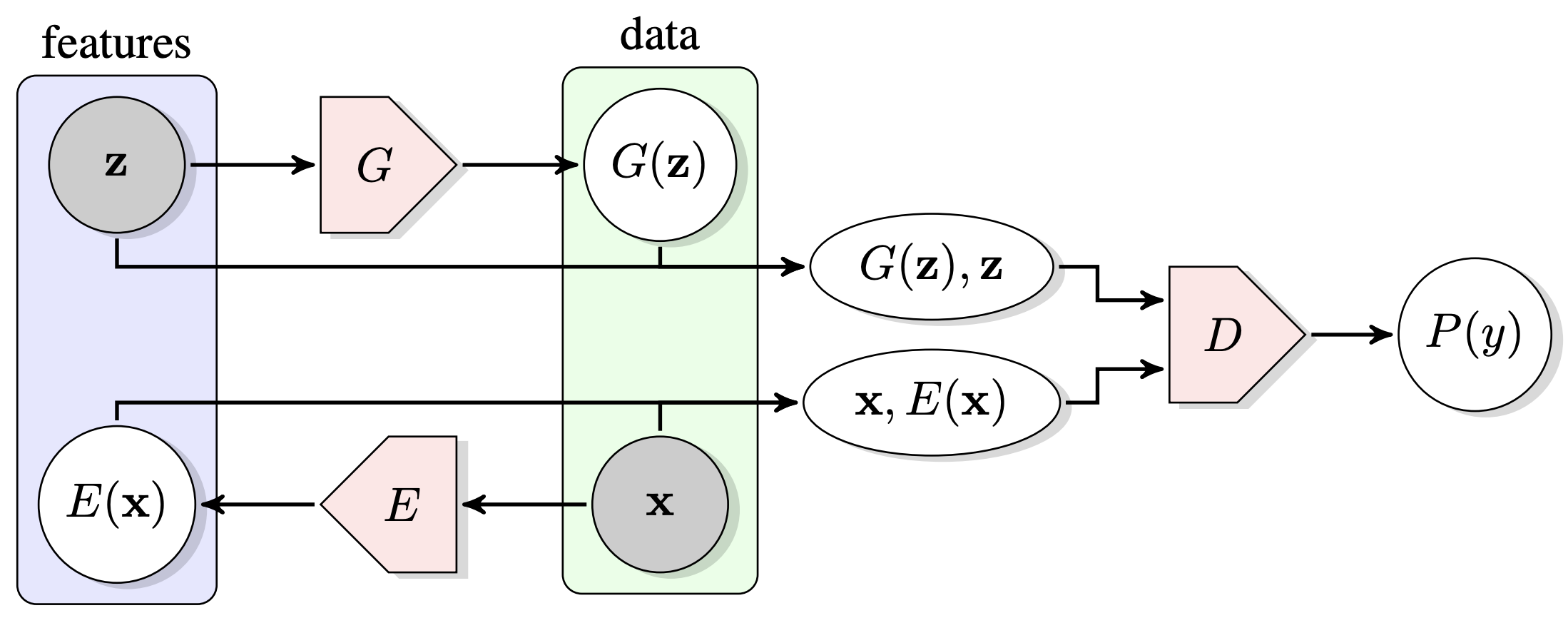}
    \caption{The architecture of BiGAN~\cite{donahue2016adversarial} and ALI~\cite{dumoulin2016adversarially}. This framework jointly models the generation and encoding. Taken from~\cite{donahue2016adversarial}}
    \label{fig:BiGAN}
    \vspace{-0.2cm}
\end{figure}
}

However, AAE still preserves the reconstruction error, which contradicts GAN's core idea. Based on AAE, BiGAN~\cite{donahue2016adversarial} and ALI~\cite{dumoulin2016adversarially} argue to embrace adversarial learning without reservation and put forward a new framework. Given an actual sample $x$

\begin{itemize}
    \item Generator $G$: the generator here virtually acts as the decoder, generates fake samples $x^\prime=G(z)$ by $z$ from a prior latent distribution (e.g. [uniform(-1,1)]$^d$, d refers to dimension).
    \item Encoder $E$: a newly added component, mapping real sample $x$ to representation $z^\prime=E(x)$. This is also exactly what we want to train.
    \item Discriminator $D$: given two inputs [$z$, $G(z)$] and [$E(x)$, $x$], decide which one is from the real sample distribution.
\end{itemize}

It is easy to see that their training goal is $E = G^{-1}$. In other words, encoder $E$ should learn to "convert" generator $G$. This goal could be rewritten as a $l_0$ loss for autoencoder~\cite{donahue2016adversarial}, but it is not the same as a traditional autoencoder because the distribution does not make any assumption about the data itself. The distribution is shaped by the discriminator, which captures the semantic-level difference. Based on BiGAN and ALI, later studies~\cite{chongxuan2017triple,donahue2019large} discover that GAN with deeper and larger networks and modified architectures can produce even better results on downstream task.

\subsection{Recover with Partial Input}
As we mentioned above, GAN's architecture is not born for representation learning, and modification is needed to apply its framework. While BiGAN and ALI choose to extract the implicit distribution directly, some other methods such as colorization~\cite{zhang2016colorful,zhang2017split,larsson2017colorization,larsson2016learning}, inpainting~\cite{iizuka2017globally,pathak2016context} and super-resolution~\cite{ledig2017photo} apply the adversarial learning via in a different way. Instead of asking models to reconstruct the whole input, they provide models with partial input and ask them to recover the rest parts. This is similar to denoising autoencoder (DAE) such as BERT's family in natural language processing but conducted in an adversarial manner.

Colorization is first proposed by ~\cite{zhang2016colorful}. The problem can be described as given one color channel $L$ in an image and predicting the value of two other channels $A$, $B$. The encoder and decoder networks can be set to any form of convolutional neural network. Interestingly, to avoid the uncertainty brought by traditional generative methods such as VAE, the author transforms the generation task into a classification one. The first figure out the common locating area of $(A, B)$ and then split it into 313 categories. The classification is performed through a softmax layer with hyper-parameter $T$ as an adjustment. Based on ~\cite{zhang2016colorful}, a range of colorization-based representation methods~\cite{zhang2017split,larsson2017colorization,larsson2016learning} are proposed to benefit downstream tasks.

\begin{figure}[t!]
    \centering
    \includegraphics[width=.48\textwidth]{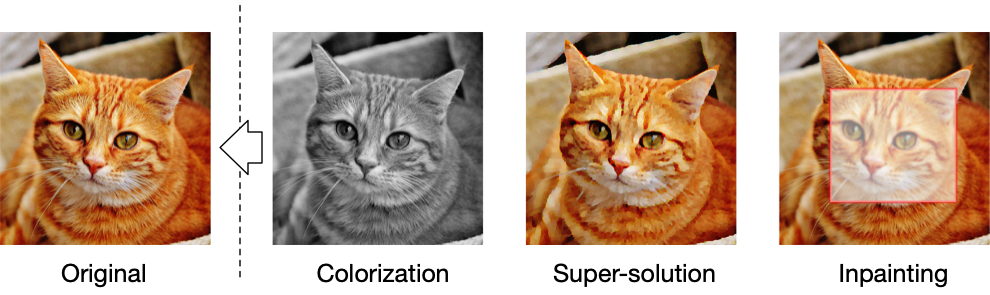}
    \caption{Illustration of typical "recovering with partial input" methods: colorization, inpainting and super-resolution. Given the original input on the left, models are asked to recover it with different partial inputs given on the right.}
    \label{fig:partial_input}
    \vspace{-0.2cm}
\end{figure}

Inpainting~\cite{iizuka2017globally,pathak2016context} is more straight forward. We will ask the model to predict an arbitrary part of an image given the rest of it. Then a discriminator is employed to distinguish the inpainted image from the original one. Super-resolution method SRGAN~\cite{ledig2017photo} follows the same idea to recover high-resolution images from blurred low-resolution ones in the adversarial setting.

\subsection{Pre-trained Language Model}
For a long time, the pre-trained language model (PTM) focuses on maximum likelihood estimation based pretext task because discriminative objectives are thought to be helpless due to languages' vibrant patterns. However, recently some work shows excellent performance and sheds light on contrastive objectives' potential in PTM.

\begin{figure}[h!]
    \centering
    \includegraphics[width=.48\textwidth]{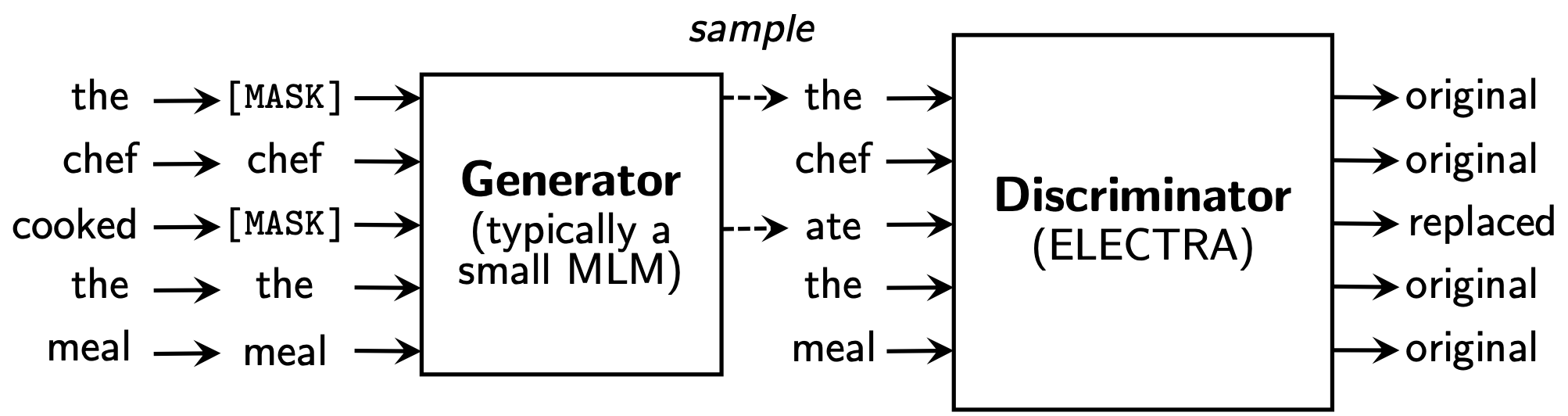}
    \caption{The architecture of ELECTRA~\cite{clark2020electra}. It follows GAN's framework but uses a two-stage training paradigm to avoid using policy gradient. The MLM is Masked Language Model.}
    \label{fig:ELECTRA}
    \vspace{-0.2cm}
\end{figure}

The pioneering work is ELECTRA~\cite{clark2020electra}, surpassing BERT given at the same computation budget. ELECTRA proposes Replaced Token Detection (RTD) and leverages GAN's structure to pre-train a language model. In this setting, the generator $G$ is a small Masked Language Model (MLM), which replaces masked tokens in a sentence to words. The discriminator $D$ is asked to predict which words are replaced. Notice that \textit{replaced} means not the same with original unmasked inputs. The training is conducted in two stages:

\begin{enumerate}
    \item Warm-up the generator: train the $G$ with MLM pretext task $\mathcal{L_{\rm MLM}(\boldsymbol{x}, \theta_{\rm G})}$ for some steps to warm up the parameters.
    \item Trained with the discriminator: $D$'s parameters is initialized with $G$'s and then trained with the discriminative objective $\mathcal{L_{\rm Disc}(\boldsymbol{x}, \theta_{\rm D})}$ (a cross-entropy loss). During this period, $G$'s parameter is frozen.
\end{enumerate}

\noindent The final objective could be written as

\beq{
    \min\limits_{\theta_{\rm G},\theta_{\rm D}}\sum\limits_{\boldsymbol{x}\in\mathcal{X}}\mathcal{L_{\rm MLM}(\boldsymbol{x}, \theta_{\rm G})}+\lambda\mathcal{L_{\rm Disc}(\boldsymbol{x}, \theta_{\rm D})}
}

Though ELECTRA is structured as GAN, it is not trained in the GAN setting. Compared to image data, which is continuous, word tokens are discrete, which stops the gradient backpropagation. A possible substitution is to leverage policy gradient, but ELECTRA experiments show that performance is slightly lower. Theoretically speaking, $\mathcal{L_{\rm Disc}(\boldsymbol{x}, \theta_{\rm D})}$ is actually turning the conventional $k$-class softmax classification into a binary classification. This substantially saves the computation effort but may somehow harm the representation quality due to the early degeneration of embedding space. In summary, ELECTRA is still an inspiring pioneer work in leveraging discriminative objectives.

At the same time, WKLM~\cite{xiong2019pretrained} proposes to perform RTD at the entity-level. For entities in Wikipedia paragraphs, WKLM replaced them with similar entities and trained the language model to distinguish them in a similar discriminative objective as ELECTRA, performing exceptionally well in downstream tasks like question answering. Similar work is REALM~\cite{guu2020realm}, which conducts higher article-level retrieval augmentation to the language model. However, REALM is not using the discriminative objective.

\subsection{Graph Learning}
There are also attempts to utilize adversarial learning (~\cite{dai2018adversarial,wang2018graphgan,ding2018semi}). Interestingly, their ideas are quite different from each other.

The most natural idea is to follow BiGAN~\cite{donahue2016adversarial} and ALI~\cite{dumoulin2016adversarially}'s a practice that asks discriminator to distinguish representation from generated and prior distribution. Adversarial Network Embedding (ANE)~\cite{dai2018adversarial} designs a generator $G$ that is updated in two stages: 1) $G$ encodes sampled graph into target embedding and computes traditional NCE with a context encoder $F$ like Skip-gram, 2) discriminator $D$ is asked to distinguish embedding from $G$ and a sampled one from a prior distribution. The optimized objective is a sum of the above two objectives, and the generator $G$ could yield better node representation for the classification task.

\hide{
\begin{figure}[h!]
    \centering
    \includegraphics[width=.48\textwidth]{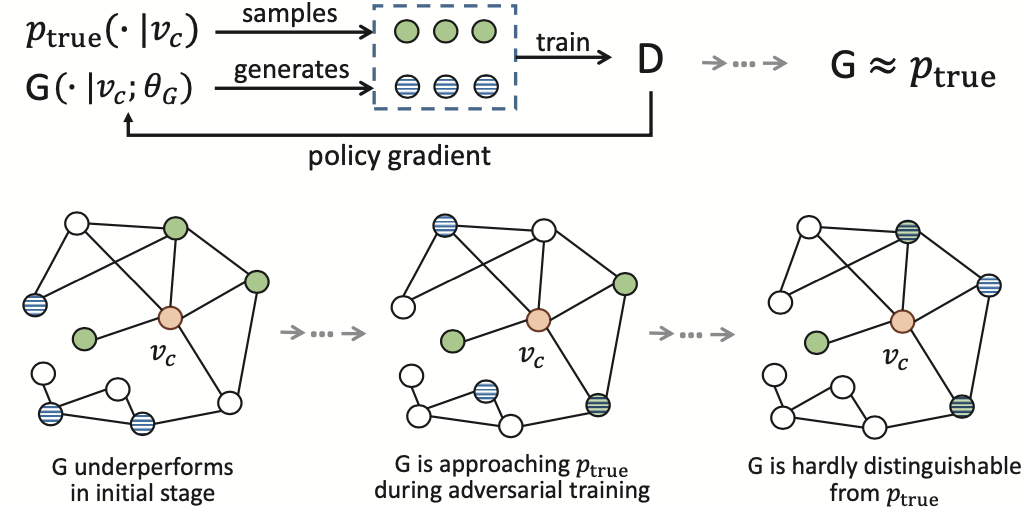}
    \caption{The architecture of GraphGAN.}
    \label{fig:GraphGAN}
    \vspace{-0.2cm}
\end{figure}
}
GraphGAN~\cite{wang2018graphgan} considers to model the link prediction task and follow the original GAN style discriminative objective to distinguish directly at node-level rather than representation-level. The model first selects nodes from the target node's subgraph $v_c$ according to embedding encoded by the generator $G$. Then some neighbor nodes to $v_c$ selected from the subgraph, together with those selected by $G$, are put into a binary classifier $D$ to decide whether they are linked to $v_c$. Because this framework involves a discrete selection procedure, while gradient descents could update the discriminator $D$, the generator $G$ is updated via policy gradients.

\begin{figure}[h!]
    \centering
    \includegraphics[width=.35\textwidth]{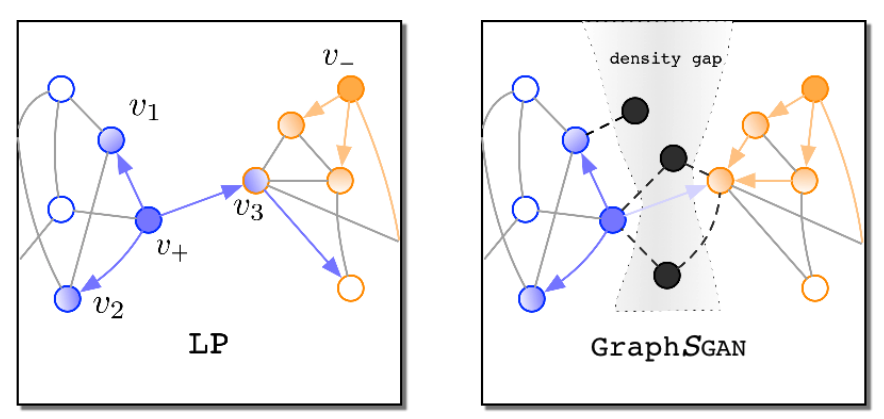}
    \caption{The architecture of GraphSGAN~\cite{ding2018semi}, which investigates density gaps in embedding space for classification problems. Taken from~\cite{ding2018semi}}
    \label{fig:GraphGAN}
    \vspace{-0.2cm}
\end{figure}
GraphSGAN~\cite{ding2018semi} applies the adversarial method in semi-supervised graph learning with the motivation that marginal nodes cause most classification errors in the graph. Consider samples in the same category; they are usually \textit{clustered} in the embedding space. Between clusters, there are \textit{density gaps} where few samples exist. The author provides rigorous mathematical proof that we can perform complete classification theoretically if we generate enough fake samples in density gaps. GraphSGAN leverages a generator $G$ to generate fake nodes in density gaps during the training and asks the discriminator $D$ to classify nodes into their original categories and a category for those fake ones. In the test period, fake samples are removed, and classification results on original categories could be improved substantially.

\subsection{Domain Adaptation and Multi-modality Representation}
Essentially, the discriminator in adversarial learning serves to match the discrepancy between latent representation distribution and data distribution. This function naturally relates to domain adaptation and multi-modality representation problems, aiming to align different representation distribution. ~\cite{ajakan2014domain,ganin2016domain,shen2017adversarial,alam2018domain} studies how GAN can help on domain adaptation. ~\cite{cai2017kbgan,wang2018incorporating} leverage adversarial sampling to improve the negative samples' quality. For multi-modality representation, ~\cite{zhu2017toward}'s image to image translation,~\cite{shen2017style}'s text style transfer,~\cite{conneau2017word}'s word to word translation and ~\cite{sarafianos2019adversarial} image to text translation show great power of adversarial representation learning.

\subsection{Pros and Cons}
Generative-contrastive (adversarial) self-supervised learning is particularly successful in image generation, transformation and manipulation, but there are also some challenges for its future development:

\begin{itemize}
    \item \textbf{Limited applications in NLP and graph.} Due to the discrete nature of languages and graphs, the adversarial methods do not perform as well as they do in computer vision. Furthermore, GAN-based language generation has been found to be much worse than unidirectional language models such as GPTs.
    \item \textbf{Easy to collapse.} It is also notorious that adversarial models are prone to collapse during the training, with numerous techniques developed to stabilize its training, such as spectral normalization~\cite{miyato2018spectral}, W-GAN~\cite{arjovsky2017wasserstein} and so on.
    \item \textbf{Not for feature extraction.} Although works such as BiGAN~\cite{donahue2016adversarial} and BigBiGAN~\cite{donahue2019large} have explored some ways to leverage GAN's learned latent representation and achieve good performance, contrastive learning has soon outperformed them with fewer parameters. 
\end{itemize}

Despite the challenges, however, it is still promising because it overcomes some inherent deficits of the point-wise generative objective. Maybe we still need to wait for a better future implementation of this idea.

%% file: 6-theory.tex
\section{Theory behind Self-supervised Learning} \label{section:theory}
In last three sections, we introduces a number of empirical works for self-supervised learning. However, we are also curious about their theoretical foundations. In this part, we will provide some theoretical insights on self-supervised learning's success from different perspectives.

\hide{Generally speaking, representation learning can be viewed as a process to optimize the difference between two distributions. In this part, we show the optimization target of part of the models, as mentioned earlier, from the perspective of probability and figure out the implicit links.}

\subsection{GAN}
\subsubsection{Divergence Matching}
As generative models, GANs\cite{goodfellow2014generative} pays attention to the difference between real data distribution $P_{data}(x)$ and generated data distribution $P_{G}(x;\theta)$:
\begin{equation}
    \label{gan_init}
    \begin{split}
        \theta^* 
        &= \mathop{\arg\max}_{\theta} \Pi_{i=1}^{m}P_{G}(x^i;\theta) \\
        &\approx \mathop{\arg\min}_{\theta} {\rm KL}(P_{data}(x) || P_{G}(x;\theta))      
    \end{split}
\end{equation}

f-GAN\cite{nowozin2016f} shows that the generative-adversarial approach is a special case of an exsiting more general variational divergence estimation problem, and uses f-divergence to train the generative models. f-divergence reflects the difference of two distributions $P$ and $Q$:
\begin{equation}
    \begin{split}
        \mathcal{D}_{f}(P||Q) 
        &= \mathbb{E}_{q(x)}[f(\frac{p(x)}{q(x)})] \\
        &= \max_{T}(\mathbb{E}_{x \sim p(x)}[T(x)] - \mathbb{E}_{x \sim q(x)}[g(T(x))])
    \end{split}
    \label{fgan}
\end{equation}
Replace KL-divergence in (\ref{gan_init}) with Jensen-Shannon(JS) divergence $JS=\frac{1}{2}[\mathbb{E}_{p(x)}\log\frac{2p(x)}{p(x)+q(x)} +\mathbb{E}_{q(x)}\log\frac{2q(x)}{p(x)+q(x)}]$ and calculate the replaced one with (\ref{fgan}), the optimization target of the minmax GAN is achieved.
\begin{equation}
    \label{gan_target}
    \min_{G} \max_{D}(\mathbb{E}_{P_{data}(x)}[\log D(x)] + \mathbb{E}_{P_{G}(x;\theta)}[\log(1-D(x))])
\end{equation}
Different divergence functions leads to different GAN variants. \cite{nowozin2016f} also discusses the effects of various choices of divergence functions.

\subsubsection{Disentangled Representation}
An important drawback of supervised learning is that it easily get trapped into spurious information. A famous example is that supervised neural networks learn to distinguish dogs and wolves by whether they are in the grass or snow~\cite{ribeiro2016should}, which means the supervised models do not learn the disentangled representations of the grass and the dog, which should be mutual independent.

As an alternative, GAN show its superior potential in learning disentangled features empirically and theoretically. InfoGAN~\cite{chen2016infogan} first proposes to learn disentangled representation with DCGAN. Conventionally, we sample white noise from a uniform or Gaussian distribution as input to generator of GAN. However, this white noise does not make any sense to the characteristics of the image we generated. We assume that there should be a latent code $c$ whose dimensions represent different characteristics of the image respectively (such as rotation degree and width). We will learn this $c$ jointly in the discrimination period by the discriminator $D$, and maximize $c$'s mutual information $I(c;x)$ with the image $x=G(z,c)$, where $G$ refers to the generator (actually the decoder).

Since mutual information is notoriously hard to compute, the authors leverage the variational inference approcach to estimates its lower bound $L_I(c,x)$, and the final objective for InfoGAN is modified as:

\beq{
    \min\limits_G\max\limits_D V_I(D,G)=V(D,G)-\lambda L_I(c;G(z,c))
}

\noindent Experiments show that InfoGAN surely learns a good disentangled representation on MNIST. This further encourage researchers to identify whether the modular structures for generation inner the GAN could be disentangled and independent with each others. GAN dissection~\cite{bau2018gan} is a pioneer work in applying causal analysis into understading GAN. They identify the correlations between channels in the convolutional layers and objects in the generated images, and examine whether they are causally-related with the output. ~\cite{besserve2018counterfactuals} takes another step to examine these channels' conditional independence via rigorous counterfactual interventions over them. Results indicate that in BigGAN researchers are able to disentangle backgrounds and objects, such as replacing the background of a cock from the bare soil with the grassland. 

These work indicates the ability of GAN to learn disentangled features and other self-supervised learning methods are likely to be capable too.

\begin{figure}[t!]
    \centering
    \includegraphics[width=.38\textwidth]{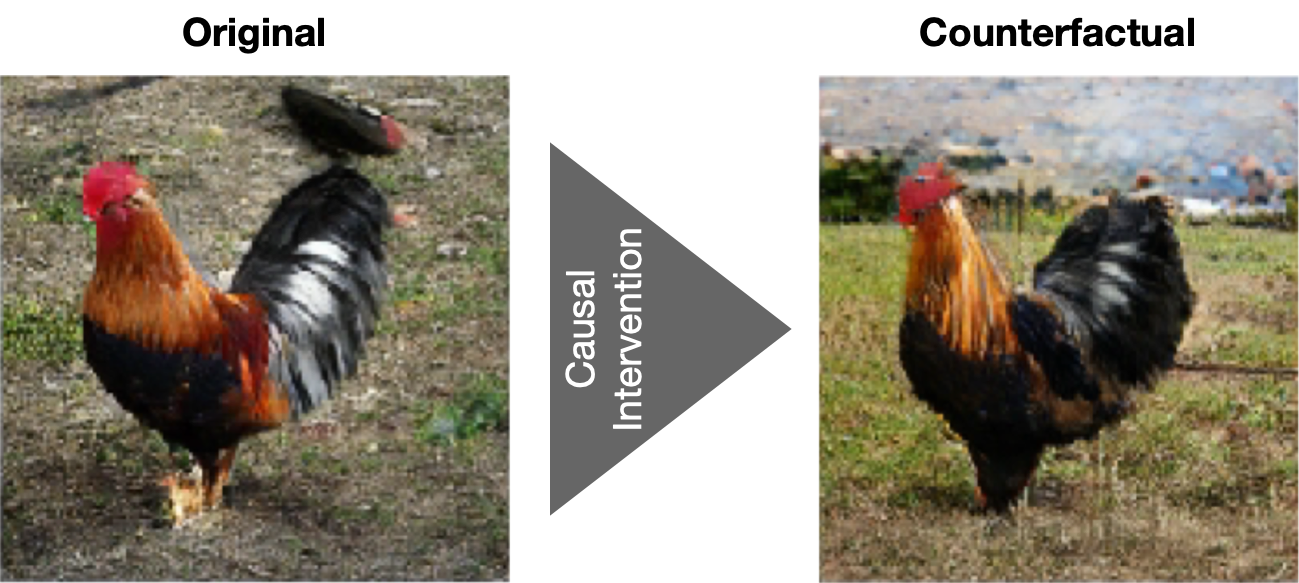}
    \caption{GAN is able to learn disentangled features and encode them in its modular structure. In~\cite{besserve2018counterfactuals}, researchers show that in GAN the features of cock is disentangled with the features of background. Repainted from~\cite{besserve2018counterfactuals}}
    \label{fig:disentangle}
    \vspace{-0.2cm}
\end{figure}

\subsection{Maximizing Lower Bound}

\subsubsection{Evidence Lower Bound}
VAE (variational auto-encoder) learns the representation through learning a distribution $q_{\phi}(z|x)$ to approximate the posteriori distribution $p_{\theta}(z|x)$, 
\begin{equation}
\begin{split}
    {\rm KL}(q_{\phi}(z|x)||p_{\theta}(z|x)) = -{\rm ELBO}(\theta;\phi;x) + \log p_{\theta}(x) 
\end{split}
\end{equation}
\begin{equation}
    {\rm ELBO}= E_{q_{\phi}(z|x)}[\log q_\phi(z|x)] - E_{p_{\theta}}[\log p_{\theta}(z,x)]    
\end{equation}

\noindent where ELBO (Evidence Lower Bound Objective) is the lower bound of the optimization target ${\rm KL}(q_{\phi}(z|x)||p_{\theta}(z|x))$. VAE maximizes the $ELBO$ to minimize the difference between $q_{\phi}(z|x)$ and $p_{\theta}(z|x)$. 
\begin{equation}
    {\rm ELBO}(\theta;\phi;x) = -{\rm KL}(q_{\phi(z|x)||p_{\theta}(z)}) + E_{q_{\phi}}[\log p_{\theta}(x|z)]
\end{equation}

\noindent where ${\rm KL}(q_{\phi(z|x)||p_{\theta}(z)})$ is the regularization loss to approximate the Gaussian Distribution and $E_{q_{\phi}}[\log p_{\theta}(x|z)]$ is the reconstruction loss.

\subsubsection{Mutual Information}
    Most of current contrastive learning methods aim to maximize the MI(Mutual Information) of the input and its representation with joint density $p(x,y)$ and marginal densities $p(x)$ and $p(y)$:
\begin{equation}
    \begin{split}
        I(X,Y) &= \mathbb{E}_{p(x,y)}[\log\frac{p(x,y)}{p(y)p(x)}] \\ 
        &= {\rm KL}(p(x,y)|p(x)p(y))
    \end{split}
\end{equation}

Deep Infomax\cite{hjelm2018learning}w maximizes the MI of local and global features and replaces KL-divergence with JS-divergence, which is similar to GAN mentioned above. Therefore the optimization target of Deep Infomax becomes:
\begin{equation}
    \max_{T}(\mathbb{E}_{p(x,y)} [\log (T(x,y)] + \mathbb{E}_{p(x)p(y)}[\log(1-T(x,y))])
\end{equation}

The form of the objective optimization function is similar to (\ref{gan_target}), except that the data distribution becomes the global and local feature distributions. From a probability point of view, GAN and DeepInfoMax are derived from the same process but for a different learning target. The encoder in GAN, to an extent, works the same as the encoder in representation learning models. The idea of generative-contrastive learning deserves to be used in self-learning areas.  

Instance Discrimination\cite{wu2018unsupervised}\cite{oord2018representation} directly optimizes the proportion of gap of positive pairs and negative pairs. One of the commonly used estimators is InfoNCE\cite{oord2018representation}:
\begin{equation}
    \label{infonce}
    \begin{split}
    \mathcal{L} 
    &= \mathbb{E}_{X}[-\log \frac{\exp(x \cdot y / \tau)}{\Sigma_{i=0}^{K}\exp(x \cdot x^-/\tau) + \exp(x \cdot y / \tau)}]\\
    &= \mathbb{E}_{X}[-\log\frac{p(x|y)/p(x)}{p(x|y)/p(x) + \Sigma_{x^- \in \mathbb{X_{\neg}}} p(x^-|y)/p(x^-) }] \\ 
    &\approx \mathbb{E}_{X}\log[1+\frac{p(x)}{p(x|y)} (N-1) \mathbb{E}_{x^-}\frac{p(x^-|y)}{p(x^-)}] \\
    & \geq \mathbb{E}_{X} \log[\frac{p(x)}{p(x|y)}N]    \\
    &= -I(y, x) + \log(N)   
    \end{split}
\end{equation}

Therefore the MI $I(x,y) \geq \log(N) - \mathcal{L}$. The approximation becomes increasingly accurate, and $I(x,y)$ also increases as N grows. This implies that it is useful to use large negative samples(large values of N). But \cite{arora2019theoretical} has demonstrated that increasing the number of negative samples does not necessarily help. Negative sampling remains a key challenge to study.

Though maximizing ELBO and MI has been achieved to obtain the state-of-art result in self-supervised representation learning, it is demonstrated that MI and ELBO are loosely connected with the downstream task performance\cite{kingma2013auto}\cite{tschannen2019mutual}. Maximizing the lower bound(MI and ELBO) is not sufficient to learn useful representations. On the one hand, looser bounds often yield better test accuracy in downstream tasks. On the other hand, achieving the same lower bound value can lead to vastly different representations and performance on downstream tasks, which indicates that it does not necessarily capture useful information of data\cite{alemi2017fixing}\cite{tschannen2018recent}\cite{blau2019rethinking}. There is a non-trivial interaction between the representation encoder, critic, and loss function\cite{tschannen2019mutual}.

MI maximization can also be analyzed from the metric learning view. \cite{tschannen2019mutual} provides some insight by connecting InfoNCE to the triplet (k-plet) loss in deep learning community. The InfoNCE (\ref{infonce}) can be rewriten as follows:
\begin{equation}
    \begin{split}
        I_{NCE} 
        &= \mathbb{E}[\frac{1}{k} \Sigma_{i=1}^{k}\log \frac{e^{f(x_i, y_i)}}{\frac{1}{k} \Sigma_{j=1}^{k} e^{f(x_i, y_j)} }] \\
        &= \log k - \mathbb{E}[\frac{1}{k} \Sigma_{i=1}^{k} \log(1+\Sigma_{j \neq i}^{k}e^{f(x_i,y_j)-f(x_i, y_i)})]
    \end{split}
\end{equation}

In particular $f$ is contrained to the form $f(x,y)=\phi(x)^T \phi(y)$ for a certain function $\phi$. Then the InfoNCE is corresponding to the expectation of the \textit{multi-class k-pair} loss:
\begin{equation}
    \label{k_pair}
    L_{k-pair}(\phi) = \frac{1}{k} \Sigma_{i=1}^{k}\log(1+\Sigma_{j \neq i} e^{\phi(x_i)^T \phi(y_j) - \phi(x_i)^T\phi(y_i)}) 
\end{equation}
In metric learning, the encoder is shared across views($\phi(x)$ and $\phi(y)$) and the critic function $f(x,y) = \phi(x)^T\phi(y)$ is symmetric, while the MI maximization is not constrained by these conditions. (\ref{k_pair}) can be viewed as learning encoders with a parameter-less inner product.

\subsection{Contrastive Self-supervised Representation Learning}
\subsubsection{Relationship with Supervised Learning}
Self-supervised learning, as is indicated literally, follows the supervised learning pattern. Empirical evidence shows that contrastive learning for pre-training is especially effective for downstream classification tasks (while this improvement is not obvious on many generation tasks). We want to know how contrastive pre-training benefits supervised learning, especially on whether self-supervised learning could learn more, at least for accuracy, than supervised learning does.

\begin{figure}[h]
    \centering
    \includegraphics[width=.47\textwidth]{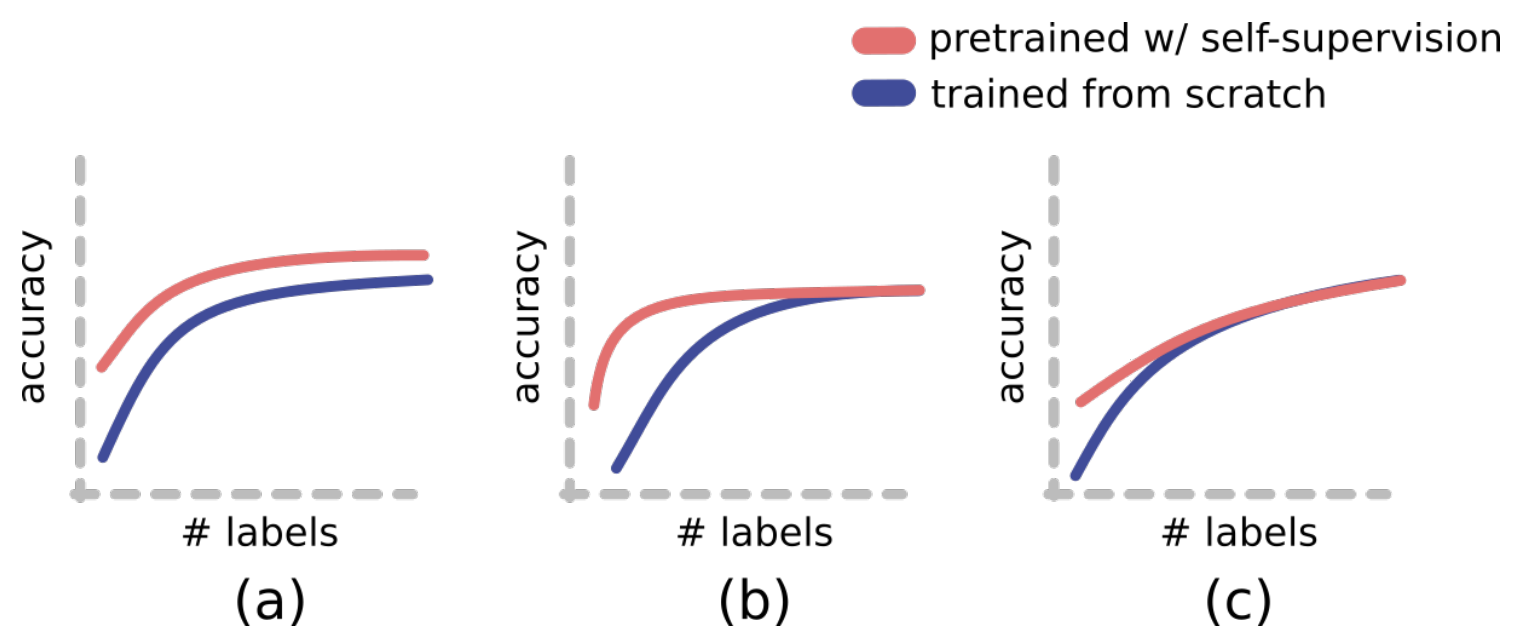}
    \caption{Three possible assumptions about supervised pre-training v.s. supervised training from scratch. Taken from~\cite{newell2020useful}}
    \label{fig:useful}
    \vspace{-0.2cm}
\end{figure}

Newell et al.~\cite{newell2020useful} examine the three possible assumptions in Figure \ref{fig:useful} that pre-training: (a) always provides an improvement, (b) reaches higher accuracy with fewer labels but plateaus to the same accuracy as baseline, (c) converges to baseline performance before accuracy plateaus. They conducts experiment on synthetic COCO by rendering which can provide as many labels as possible and discover that self-supervised pre-training follows the patterns in (c), indicating that self-supervised learning cannot learn more than supervised learning, but can make it with few labels.

Although self-supervised learning cannot help on improving accuracy, there are many other aspects it can learn more, such as model robustness and stability. Hendrycks et al.~\cite{hendrycks2019using} discovers that self-supervised trained neural networks are much robust to adversarial examples, label corruption, and common input corruptions. What' more, it greatly benefits out-of-distribution
detection on difficult, near-distribution outliers, so much so that it exceeds the performance of fully supervised methods.

\subsubsection{Understand Contrastive Loss} \label{sec:understand_contrastive}
In~\cite{wang2020understanding}, Wang et al. conduct interesting theoretical analysis on functions of contrastive loss, and split it into two terms:

\begin{equation}
    \begin{split}
        \mathcal{L}_{\rm contrast}
        &=\mathbb{E}[-\mathop{\log}\frac{e^{f_x^Tf_y/\tau}}{e^{f_x^Tf_y/\tau}+\sum_ie^{f_x^Tf_{y_i^-}/\tau}}]
        \\&=\underbrace{\mathbb{E}[-f_x^Tf_y/\tau]}_{\rm alignment} + \underbrace{\mathbb{E}[\mathop{\log}(e^{f_x^Tf_y/\tau}+\sum_ie^{f_x^Tf_{y_i^-}/\tau})]}_{\rm uniformity}
    \end{split}
    \label{eqa:contrastive_loss}
\end{equation}

\noindent where the first term aims at ``alignment'' and the second aims at ``uniformity'' of sample vectors on a sphere given the normalization condition. Experiments show that these two terms have a large agreement with downstream tasks. In addition, the authors explore to directly optimize \textit{alignment} and \textit{uniformity} loss as:

\begin{equation}
    \begin{split}
        \mathcal{L}_{\rm align}(f;\alpha)\triangleq\mathbb{E}_{(x,y)\sim p_{\rm pos}}[||f(x)-f(y)||_2^\alpha] \\
        \mathcal{L}_{\rm uniform}(f;t)\triangleq\mathop{\log}\mathbb{E}_{x,y\sim p_{\rm data}}[e^{-t||f(x)-f(y)||_2^2}]
    \end{split}
    \label{eqa:direct_optimize}
\end{equation}

\noindent in a joint additive form. They conduct experiments in wide range of scenarios including using CNN or RNN in computer vision or natural language processing tasks, and discover that direct optimization is consistently better than contrastive loss. Besides, both alignment and uniformity are necessary for a good representation. When one of the weights for these two losses is too large, the representation would collapse.

However, it is doubtful that whether alignment and uniformity are necessarily in the form of upper two losses, because in BYOL~\cite{grill2020bootstrap}, the authors display a framework without direct negative sampling but outperform all previous contrastive learning pre-training. This illustrates us that we may still achieve uniformity via other techniques such as exponential moving average, batch normalization, regularization and random initialization.

\subsubsection{Generalization}
It seems intuitive that minimizing the aforementioned loss functions should lead the representations better to capture the "similarity" between different entities, but it is unclear why the learned representations should also lead to better performance on downstream tasks, such as linear classification tasks. Intuitively, a self-supervised representation learning framework must capture the feature in unlabelled data and the similarity with semantic information that is implicitly present in downstream tasks. \cite{arora2019theoretical} proposed a conceptual framework to analyze contrastive learning on average classification tasks.

Contrastive learning assumes that similar data pair $(x, x^+)$ comes from a distribution $\mathcal{D}_{sim}$ and negative sample $(x_1^-, ... ,x_k^-)$ from a dstribution $\mathcal{D}_{neg}$ that is presumably unrelated to $x$. Under the hypothesis that semantically similar points are sampled from the same latent class, the unsupervised loss can be expressed as:
\begin{equation}
    \label{unp_loss}
    \mathcal{L}_{un}(f)=\mathbb{E}_{ \begin{subarray}{l}
        x^+ \sim \mathcal{D}_{sim} \\
         x^- \sim \mathcal{D}_{neg} 
    \end{subarray}}
    [l (\{ f(x)^T (f(x^+) - f(x^-))  \})]     
\end{equation}

The self-supervised learning is to find a funtion $\hat{f} \in \mathop{\arg\min}_{f} \hat{L}_{un}(f)$ that minimizes the empirical unsupervised loss within the capacity of the used encoder. As negative points are sampled independently identically from the datasets, $\mathcal{L}_{un}$ can be decomposed into $\tau \mathcal{L}_{un}^{=}$ and $(1-\tau) \mathcal{L}_{un}^{\ne}$ according to the latent class the negative sample drawed from. The intraclass deviation $s(f) \geq  c'(\mathcal{L}_{un}(f) - 1)$ controls the $\mathcal{L}_{un}(f)$ and implies the unexpected loss contradictive to our optimization target, which is caused by the negative sampling strategies. Under the context of only 1 negative sample, it is proved that optimizing unsupervised loss benefits the downstream classification tasks:
\begin{equation}
    \label{unp_to_sup}
    \mathcal{L}_{sup}(\hat{f}) \leq \mathcal{L}_{sup}^{\mu}(\hat{f}) \leq \mathcal{L}_{un}^{\ne} (f) + \beta s(f) + \eta Gen_{M}     
\end{equation}
With probability at least $1 - \delta$, $f$ is the feature mapping function the encoder can capture, $Gen_{M}$ is the generalization error. When the sampled pair $M \rightarrow \inf $ and the numebr of latent class $|C| \rightarrow \inf$, $Gen_{M}$ and $\delta \rightarrow 0$. If the encoder is powerful enough and trained using suffiently large number of samples, the learned function $f$ with low $\mathcal{L}_{un}^{\ne}$ as well as low $\beta s(f)$ will have good performance on supervised tasks (low $\mathcal{L}_{sup}(\hat{f})$).

Contrastive learning also has limitations. In fact, contrastive learning does not always pick the best supervised representation function $f$. Minimizing the unsupervised loss to get low $\mathcal{L}_{sup}(\hat{f})$ does not mean that $\hat{f} \approx \tilde{f} = \mathop{\arg\min}_{f} \mathcal{L}_{sup}$ because high $\mathcal{L}_{un}^{\ne}$ and high $s(f)$ does not imply high $\mathcal{L}_{sup}$, resulting the failure of the algorithm. 

The relationship between $\mathcal{L}_{sup}(f)$ and $\mathcal{L}_{sup}(\hat{f})$ are further explored on the condition of mean classifier loss $\mathcal{L}^{\mu}_{sup}$, where $\mu$ indicates that a label $c$ only corresponds to a embedding vector $\mu_c := \mathbb{E}_{x \sim D_{c}}[f(x)]$. If there exists a functoin $f$ that has intraclass concentration in strong sense and can separate latent classes with high margin(on average) with mean classifier, then $\mathcal{L}_{sup}^{\mu}(\hat{f})$ will be low. If $f(X)$ is $\sigma^{2}-sub-Gaussian$ in every direction for every class and has maximum norm $\mathbf{R} = \max_{x\ in \mathcal{X}} ||f(x)||$, then $\mathcal{L}_{sup}^{\mu}(\hat{f})$ can be controlled by $\mathcal{L}_{sup}^{\mu}(f)$.
\begin{equation}
    \label{sup_real}
    \mathcal{L}_{sup}^{u}(\hat{f}) \leq \gamma(f) \mathcal{L}_{\gamma(f), sup}^{\mu}(f) + \beta s(f) + \eta Gen_{M} + \epsilon
\end{equation}
For all $\epsilon > 0$ and with the probability at least $1 - \delta$, $\gamma=1 + c'\mathbf{R}\sigma \sqrt{\log \frac{\mathbf{R}}{\epsilon}}$. Under the assumption and context, optimizing the unsupervised loss indeed helps pick the best downstream task supervised loss.

As in the aformentioned models\cite{he2019momentum}\cite{chen2020improved},  (\ref{sup_real}) can also be extended to more than one negative samples for every similar pair. Then average loss is 
\begin{equation}
    L_{sup}(\hat{f}) := \mathbb{E}_{\Upsilon \sim \mathcal{D}}[\mathcal{L}_{sup}(\Upsilon, \hat{f})]
\end{equation}

Besides, the general belief is that increasing the number of negative samples always helps, at the cost of increased computational costs. Noise Contrastive Estimation(NCE)\cite{gutmann2010noise} explains that increasing the number of negative samples can provably improve the variance of learning parameters. However, \cite{arora2019theoretical} argues that this does not hold for contrastive learning and shows that it can hurt performance when the negative samples exceed a threshold.

Under the assumptions, contradictive representation learning is theoretically proved to benefit the downstream classification tasks. More detailed proofs can be found in \cite{arora2019theoretical}.  This connects the "similarity" in unlabelled data with the semantic information in downstream tasks. Though the connection temporarily is only in a restricted context, more generalized research deserves exploration.

%% file: 7-discussion.tex
\section{Discussions and Future Directions} \label{section:discussion}
In this section, we will discuss several open problems and future directions in self-supervised learning for representation.

\vpara{Theoretical Foundation}
Though self-supervised learning has achieved great success, few works investigate the mechanisms behind it. In this survey, we have listed several recent works on this topic and show that theoretical analysis is significant to avoid misleading empirical conclusions.

In~\cite{arora2019theoretical}, researchers present a conceptual framework to analyze the contrastive objective's function in generalization ability. ~\cite{tschannen2019mutual} empirically proves that mutual information is only loosely related to the success of several MI-based methods, in which the sampling strategies and architecture design may count more. This type of works is crucial for self-supervised learning to form a solid foundation, and more work related to theory analysis is urgently needed. 

\vpara{Transferring to downstream tasks}
There is an essential gap between pre-training and downstream tasks. Researchers design elaborate pretext tasks to help models learn some critical features of the dataset that can transfer to other jobs, but sometimes this may fail to realize. Besides, the process of selecting pretext tasks seems to be too heuristic and tricky without patterns to follow.

A typical example is the selection of pre-training tasks in BERT and ALBERT. BERT uses Next Sentence Prediction (NSP) to enhance its ability for sentence-level understanding. However, ALBERT shows that NSP equals a naive topic model, which is far too easy for language model pre-training and even decreases BERT's performance. 

For the pre-training task selection problem, a probably exciting direction would be to design pre-training tasks for a specific downstream task automatically, just as what Neural Architecture Search~\cite{zoph2016neural} does for neural network architecture.

\vpara{Transferring across datasets}
This problem is also known as how to learn inductive biases or inductive learning. Traditionally, we split a dataset into the training used for learning the model parameters and the testing part for evaluation. An essential prerequisite for this learning paradigm is that data in the real world conform to our dataset's distribution. Nevertheless, this assumption frequently fails in experiments.

Self-supervised representation learning solves part of this problem, especially in the field of natural language processing. Vast amounts of corpora used in the language model pre-training help cover most language patterns and, therefore, contribute to the success of PTMs in various language tasks. However, this is based on the fact that text in the same language shares the same embedding space. For other tasks like machine translation and fields like graph learning where embedding spaces are different for different datasets, learning the transferable inductive biases efficiently is still an open problem.

\vpara{Exploring potential of sampling strategies}
In ~\cite{tschannen2019mutual}, the authors attribute one of the reasons for the success of mutual information-based methods to better sampling strategies. MoCo~\cite{he2019momentum}, SimCLR~\cite{chen2020simple}, and a series of other contrastive methods may also support this conclusion. They propose to leverage super large amounts of negative samples and augmented positive samples, whose effects are studied in deep metric learning. How to further release the power of sampling is still an unsolved and attractive problem.

\vpara{Early Degeneration for Contrastive Learning}
Contrastive learning methods such as MoCo~\cite{he2019momentum} and SimCLR~\cite{chen2020simple} are rapidly approaching the performance of supervised learning for computer vision. However, their incredible performances are generally limited to the classification problem. Meanwhile, the generative-contrastive method ELETRA~\cite{clark2020electra} for language model pre-training is also outperforming other generative methods on several standard NLP benchmarks with fewer model parameters. However, some remarks indicate that ELETRA's performance on language generation and neural entity extraction is not up to expectations.

Problems above are probably because the contrastive objectives often get trapped into embedding spaces' early degeneration problem, which means that the model over-fits to the discriminative pretext task too early, and therefore lost the ability to generalize. We expect that there would be techniques or new paradigms to solve the early degeneration problem while preserving contrastive learning's advantages.

%% file: 7.5-conclusion.tex
\section{Conclusion} \label{section:conclusion}
This survey comprehensively reviews the existing self-supervised representation learning approaches in natural language processing (NLP), computer vision (CV), graph learning, and beyond. Self-supervised learning is the present and future of deep learning due to its supreme ability to utilize Web-scale unlabeled data to train feature extractors and context generators efficiently. Despite the diversity of algorithms, we categorize all self-supervised methods into three classes: generative, contrastive, and generative contrastive according to their essential training objectives. We introduce typical and representative methods in each category and sub-categories. Moreover, we discuss the pros and cons of each category and their unique application scenarios. Finally, fundamental problems and future directions of self-supervised learning are listed.

%% file: 8-acknowledgement.tex
\section*{ACKNOWLEDGMENTS}

The work is supported by the
National Key R\&D Program of China (2018YFB1402600),
NSFC for Distinguished Young Scholar 
(61825602),
and NSFC (61836013).

%% file: 9-revision.tex
\section*{REVISION HISTORY}
\begin{itemize}
    \item v2/v3 (June 2020): correct several typos and mistakes.
    \item v4 (July 2020): add papers newly published; add a new theoretical analysis part for contrastive objective; add semi-supervised self-training's connection with self-supervised contrastive learning.
    \item v5 (March 2021): add papers newly published; update statistics in Fig. 2 and Fig. 7; add a motivation section to better introduce the reason of using SSL; remove the preliminary section.
\end{itemize}

%% file: bio.tex
\begin{IEEEbiography}[{\vspace{-10mm}\includegraphics[width=0.7in,height=1in,clip,keepaspectratio]{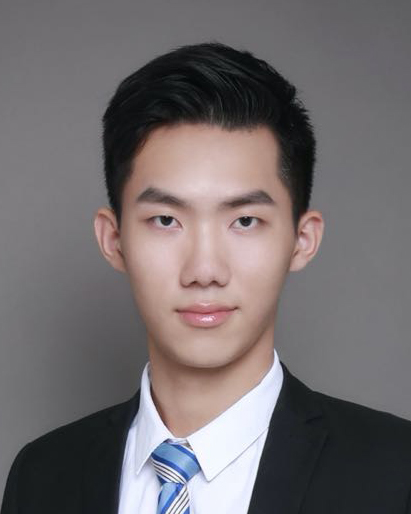}}]{Xiao Liu}
is a senior undergraduate student with the Department of Computer Science and Technology, Tsinghua University. His main research interests include data mining, machine learning and knowledge graph. He has published a paper on KDD.
\end{IEEEbiography}
\vspace{-0.76in}
\begin{IEEEbiography}[{\vspace{-10mm}\includegraphics[width=0.7in,height=1in,clip,keepaspectratio]{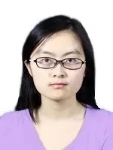}}]{Fanjin Zhang}
is a PhD candidate in the Department of Computer Science and Technology, Tsinghua University. She got her bachelor degree from the Department of Computer Science and Technology, Nanjing Unviersity. Her research interests include data mining and social network.
\end{IEEEbiography}
\vspace{-0.76in}
\begin{IEEEbiography}[{\vspace{-10mm}\includegraphics[width=0.7in,height=1in,clip,keepaspectratio]{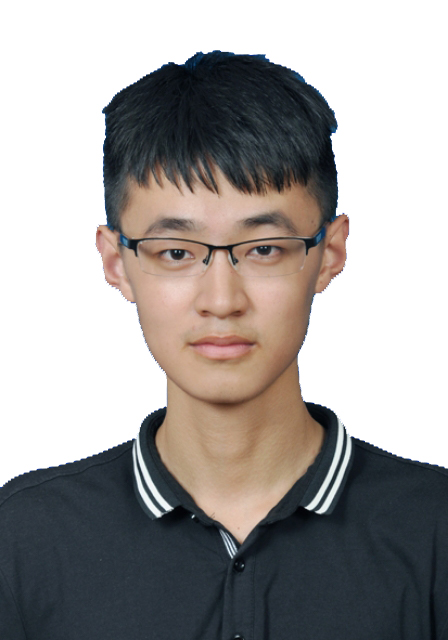}}]{Zhenyu Hou}
is an undergraduate with the department of Computer Science and Technology, Tsinghua University. His main research interests include graph representation learning and reasoning.
\end{IEEEbiography}
\vspace{-0.76in}
\begin{IEEEbiography}[{\vspace{-10mm}\includegraphics[width=0.7in,height=1in,clip,keepaspectratio]{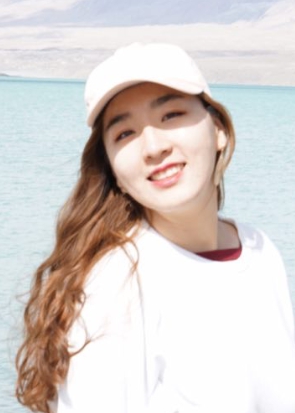}}]{Li Mian}
received bachelor degree(2020) from Department of Computer Science, Beijing Institute of Technology. She is now admitted into a graduate program in Georgia Institute of Technology. Her research interests focus on data mining, natural language processing and machine learning.
\end{IEEEbiography}
\vspace{-0.76in}
\begin{IEEEbiography}[{\vspace{-10mm}\includegraphics[width=0.7in,height=1in,clip,keepaspectratio]{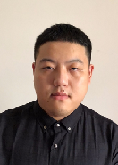}}]{Zhaoyu Wang}
is a graduate student with the Department of Computer Science and Technology of Anhui University. His research interests include data mining, natural language processing and their applications in recommender systems.
\end{IEEEbiography}
\vspace{-0.76in}
\begin{IEEEbiography}[{\vspace{-10mm}\includegraphics[width=0.7in,height=1in,clip,keepaspectratio]{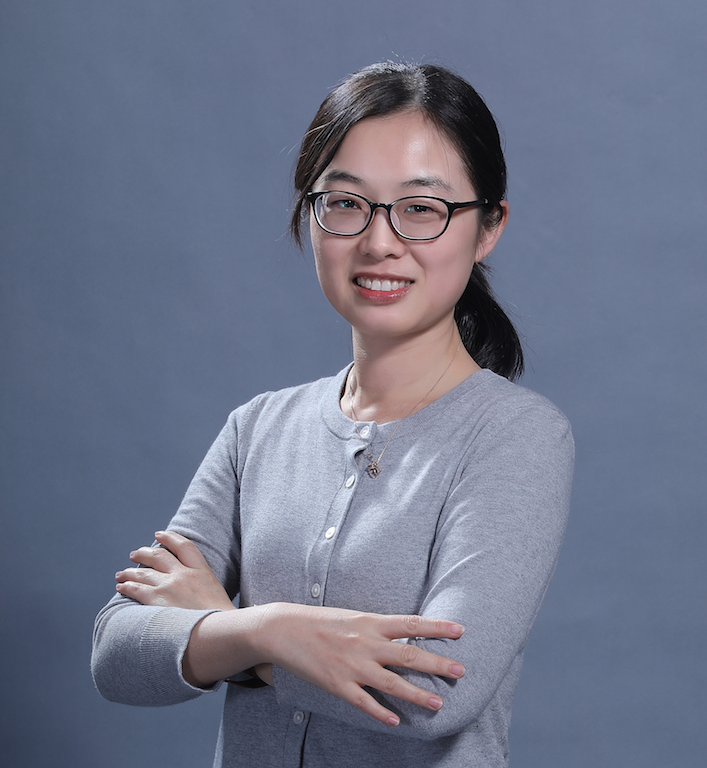}}]{Jing Zhang}
received the master and PhD
degree from the Department of Computer
Science and Technology, Tsinghua University. She is an assistant professor in Information School, Renmin University of China.
Her research interests include social network
mining and deep learning.
\end{IEEEbiography}
\vspace{-0.76in}
\begin{IEEEbiography}[{\vspace{-10mm}\includegraphics[width=0.7in,height=1in,clip,keepaspectratio]{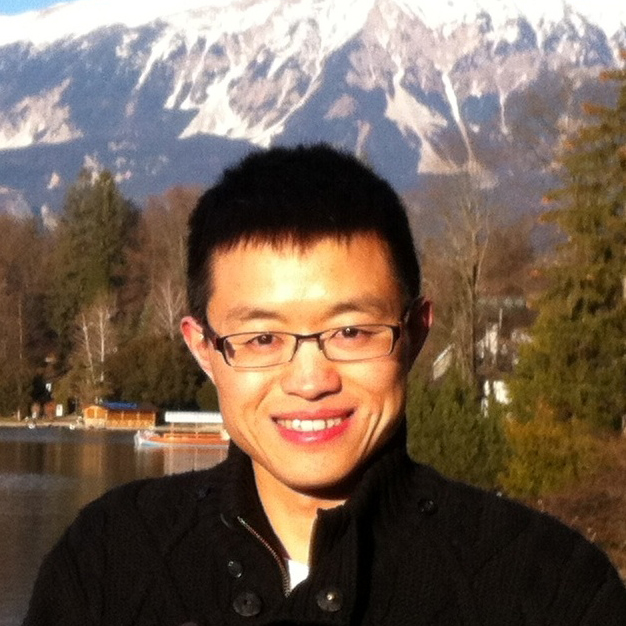}}]{Jie Tang}
received the PhD degree from Tsinghua University. He is full professor in the Department of Computer Science and Technology, Tsinghua University. His main research interests include data mining, social network, and machine learning. He has published over 200 research papers in top international journals and conferences.
\end{IEEEbiography}
\vspace{-0.76in}